\documentclass[journal,10pt]{IEEEtran}

\usepackage{multirow}

\ifCLASSOPTIONcompsoc
  \usepackage[nocompress]{cite}
\else
  \usepackage{cite}
\fi
\ifCLASSINFOpdf
   \usepackage[pdftex]{graphicx}
\else
   \usepackage[dvips]{graphicx}
\fi
\usepackage{amsmath,graphicx}
\usepackage[ruled,vlined]{algorithm2e}
\usepackage{lineno}
\usepackage{amssymb}
\usepackage{amsfonts}
\usepackage{algorithmic}
\usepackage{cite}
\usepackage{soul,color,booktabs}
\usepackage{cleveref}
\usepackage{kotex}

\usepackage{array}

\ifCLASSOPTIONcompsoc
 \usepackage[caption=false,font=footnotesize,labelfont=sf,textfont=sf]{subfig}
\else
 \usepackage[caption=false,font=footnotesize]{subfig}
\fi
\begin{document}

\title{A Multi-stage Framework with Mean Subspace Computation and Recursive Feedback for Online Unsupervised Domain Adaptation}

\author{Jihoon Moon,~\IEEEmembership{Student~Member,~IEEE}, Debasmit Das,~\IEEEmembership{Member,~IEEE} \
and C.S. George Lee,~\IEEEmembership{Fellow,~IEEE}
\thanks{Jihoon Moon 
and C.S. George Lee 
are with the Elmore Family School of Electrical and Computer Engineering, Purdue University, West Lafayette, IN 47907, USA.
Email:\{moon92, csglee\}@purdue.edu.}

\thanks{Debasmit Das (debadas@qti.qualcomm.com) is with Qualcomm Technologies Inc., San Diego, CA 92121, USA.}

\thanks{This work was supported in part by the National
Science Foundation under Grant IIS-1813935. 
Any opinion, findings,
and conclusions or recommendations expressed in this material are
those of the authors and do not necessarily reflect the views of
the National Science Foundation. }

\thanks{We also gratefully acknowledge the support of NVIDIA Corporation 
for the donation of a TITAN XP GPU used for this research.}
}

\markboth{Journal of \LaTeX\ Class Files,~Vol.~14, No.~8, August~2015}%
{Shell \MakeLowercase{\textit{et al.}}: Bare Demo of IEEEtran.cls for Computer Society Journals}

\IEEEtitleabstractindextext{
\begin{abstract}
In this paper, we address the {\em Online} Unsupervised Domain Adaptation (OUDA) problem
and propose a novel multi-stage framework to solve real-world situations 
when the target data are unlabeled and arriving online sequentially in batches. 
Most of the traditional manifold-based methods on the OUDA problem 
focus on transforming each arriving target data to the source domain 
without sufficiently considering the temporal coherency 
and accumulative statistics among the arriving target data. 
In order to project the data from the source and the target domains 
to a common subspace and manipulate the projected data in real-time,
our proposed framework institutes a novel method, 
called an Incremental Computation of Mean-Subspace (ICMS) technique,
which computes an approximation of mean-target subspace on a Grassmann manifold 
and is proven to be a close approximate to the Karcher mean. 
Furthermore, the transformation matrix computed from the mean-target subspace 
is applied to the next target data in the recursive-feedback stage, 
aligning the target data closer to the source domain. 
The computation of transformation matrix 
and the prediction of next-target subspace
leverage the performance of the recursive-feedback stage
by considering the cumulative temporal dependency 
among the flow of the target subspace on the Grassmann manifold.
The labels of the transformed target data are predicted by 
the pre-trained source classifier, then the classifier is updated 
by the transformed data and predicted labels.
Extensive experiments on six datasets were conducted to 
investigate in depth the effect and contribution 
of each stage in our proposed framework and 
its performance over previous approaches 
in terms of classification accuracy and computational speed.
In addition, the experiments on 
traditional manifold-based learning models and neural-network-based learning models 
demonstrated the applicability of our proposed framework for various types of learning models.
\end{abstract}

}

\maketitle
\IEEEdisplaynontitleabstractindextext

%
\IEEEpeerreviewmaketitle

\vspace*{0.2in}
\IEEEraisesectionheading{\section{Introduction}\label{sec:introduction}}


\IEEEPARstart{D}{omain} Adaptation (DA)~\cite{patel2015visual} 
has been a research area of growing interest 
to overcome real-world domain shift issues. 
The goal of DA is to learn a model from the source domain
with sufficient labeled data 
to maintain the performance of the learned model in the target domain,
which has a different distribution from the source domain.
The Unsupervised DA (UDA) 
problem~\cite{gong2012geodesic, zhang2017joint, fernando2013unsupervised, vascon2019unsupervised}, 
which is a branch of DA problem, 
assumes that the target data are completely unlabeled,
and the Online DA problem~\cite{gaidon2015online} tackles the DA problem 
when a learning system receives streaming target data 
in an online fashion.

Recently many studies have been conducted 
on the Online Unsupervised Domain Adaptation (OUDA) problem, 
which faces the challenges of both online DA and unsupervised DA problems.
The OUDA problem assumes that the target data are unlabeled \textit{and} 
arriving sequentially in an online fashion. 
This problem is challenging 
since it has to overcome the distribution shift 
between the source and the target domains 
while the target data is not given as an entire batch.
Throughout this paper, we use the term \textit{mini-batch} 
to indicate a bunch of consecutive target samples 
arriving at each timestep.

Early works tackled the OUDA problem 
by projecting the source and the target data to a manifold.
Bitarafan et al.~\cite{bitarafan2016incremental} 
computed a transformation matrix 
that aligned each arriving target mini-batch closer to the source domain.  
Target samples are then merged to the source domain 
once pseudo-labels of those target mini-batches are predicted.  
Hoffman et al.~\cite{hoffman2014continuous} 
adopted an additional loss term for optimization 
that minimized the difference among the target mini-batches at adjacent timesteps.
However, these approaches did not sufficiently consider 
the temporal dependency 
among the entire sequence of target mini-batches.

Majority of recent studies have tackled the OUDA problem 
by updating the model as the target mini-batch arrives. 
Some studies augmented the input data using the pseudo labels of previous target data, 
while others focused on updating the weights of the neural network 
based on the statistics (e.g., covariance, entropy etc.) of unlabeled target data.
These approaches are valid under the assumption that the target domain shifts gradually~\cite{kumar2020understanding}. 
If the source and the target distributions are significantly different, 
updating the learning system with the target data leads to catastrophic forgetting~\cite{bobu2018adapting}.

Previously, we have proposed a multi-stage framework 
for the OUDA problem~\cite{moon2020multi}, 
which computes 
the transformation matrix from target data to the source domain 
in an online fashion.
The framework also considered the accumulative statistics as well as 
temporal dependency among the entire sequence of target mini-batches.  
The proposed method incrementally computes 
the average of the low-dimensional subspaces from the target mini-batches, 
followed by the computation of a transformation matrix 
from the mean-target subspace to the source subspace.  
This transformation matrix aligns the target samples 
closer to the source domain, 
which leverages the performance of the model trained on the source domain.  

In this paper, we further extend this multi-stage OUDA framework, 
taking into account the domain shift of the evolving target domain 
as well as the shift between the source and the target domains.
In addition to computing the average subspace of target subspaces 
on a Grassmann manifold~\cite{moon2020multi}, 
we also consider the flow of the target subspaces on the Grassmann manifold.
Utilizing this flow, 
our proposed framework computes robust mean-target subspaces
and precise transformation matrices.
The next-target subspace is predicted 
by extrapolation of the geodesic from previous mean-target subspaces.
This prediction technique provides the robustness of \textit{online} domain adaptation 
for noisy target subspaces.
The recursive feedback stage leverages 
the performance of our proposed framework
by transforming the next arriving target mini-batch.
The transformed target mini-batch is represented by a subspace
that is closer to the source domain, 
hence improving the performance of \textit{online} domain adaptation.
In addition, transformation matrix from the target domain to the source domain is computed cumulatively, 
considering intermediate domains between the source and the successive target subspaces.
Moreover, the classifier in our proposed framework is adaptive, 
whereas the source classifier in our previous work~\cite{moon2020multi} 
is not changing. 
This adaptive classifier leverages the performance of online adaptation 
by including the knowledge from the arriving target data.

We also validate the efficiency of 
the incremental mean-target-subspace computation technique 
on a Grassmann manifold,
called an Incremental Computation of Mean-Subspace (ICMS)~\cite{moon2020multi}.
We have also proved that the computed mean-subspace is 
close to the Karcher mean~\cite{karcher1977riemannian} -- 
one of the most common and efficient methods related to the mean on 
manifolds -- while maintaining significantly low-computational burden 
as compared to the Karcher-mean computation. 
We have also empirically shown in experiments that ICMS is faster than 
the Karcher-mean computation.
This efficient computation of a close approximate to the Karcher mean 
makes the online feedback computation of the OUDA problem possible.  
Extensive computer simulations were performed to validate 
and analyze the proposed multi-stage OUDA framework.
In the computer simulations, we further analyzed in depth 
each stage of the proposed framework 
to investigate how those stages 
contribute to solving the OUDA problem. 

In summary, the contributions of this paper are 
efficient computation of mean-target subspace on a Grassmann manifold,
robustness on noisy online domain adaptation task using recursive feedback and next-target-subspace prediction,
consideration of cumulative temporal consistency using the flow of target domain on the Grassmann manifold,
and adaptivity of classifier that is suitable for online domain adaptation.

The remainder of this paper is organized as follows. 
Section II briefly introduces related works. 
Section III formally describes the OUDA problem and 
our proposed multi-stage OUDA framework for solving it.
Section IV describes the details of 
the proposed ICMS computation
as well as the mathematical proof of subspace convergence.
Section V describes the experimental simulations and results, 
and Section VI summarizes the findings and conclusions of the paper.

\section{Related Work}
Among many approaches proposed for the UDA problem, 
the subspace-based approaches for the UDA problem focused on 
the alignment of the projected data on a common subspace of the source and the target domains.
Long et al.~\cite{long2015learning} proposed 
the concept of Maximum Mean Discrepancy (MMD)~--~
a metric that minimized the distance between mean points 
on Reproducing Kernel Hilbert Spaces (RHKS)~\cite{berlinet2011reproducing}. 
Fernando et al.~\cite{fernando2013unsupervised} suggested the Subspace Alignment (SA) technique 
that aligned the subspaces of the source and the target domains, 
and then minimized the distance between these two domains. 

Other approaches utilized a common subspace to project the data 
from the source and the target domains, 
but they did not directly align the projected data.
Gong et al.~\cite{gong2012geodesic} proposed a Geodesic Flow Kernel (GFK) technique
that computed a transform matrix using a kernel-based method.
This transform matrix characterizes the transformation of the original data
from the target domain to the source domain on a Grassmann manifold.

Other previous work attempted to align the source and the target domains 
by directly minimizing their statistical properties. 
Sun et al.~\cite{sun2016return} adopted a Correlation Alignment (CORAL) approach 
that minimized the domain discrepancy 
directly on the original data space 
by adjusting the second-order statistics of the source and the target distributions. 
Zhang et al.~\cite{zhang2017joint} suggested 
the Joint Geometrical and Statistical Alignment (JGSA) technique ~--~ 
a combined technique of MMD and SA methods. 
Wang et al.~\cite{wang2018visual} proposed 
a Manifold Embedded Distribution Alignment (MEDA) approach
that quantitatively evaluated the marginal and conditional distributions in domain adaptation.
Vascon et al.~\cite{vascon2019unsupervised} 
formulated the UDA problem as 
a semi-supervised incremental learning problem using the Game Theory 
and suggested a Graph Transduction for Domain Adaptation (GTDA) method. 
The GTDA method obtained the optimal classification result on the target domain
using Nash equilibrium~\cite{nash1950equilibrium}.
Wulfmeier et al.~\cite{wulfmeier2017addressing} adopted 
Generative Adversarial Networks (GANs)~\cite{goodfellow2014generative} 
to align the features across domains.

Recently, more studies on the OUDA problem have emerged.
Wulfmeier et al.~\cite{wulfmeier2018incremental} 
extended their previous work~\cite{wulfmeier2017addressing} 
on the UDA problem to the online case using a GAN-based approach. 
Unfortunately, their approach was not applicable to real-time situations
since it requires the training stage of the target data.
Bitarafan et al.~\cite{bitarafan2016incremental} proposed 
an Incremental Evolving Domain Adaptation (EDA) technique 
that consisted of the target data transformation using GFK 
followed by the source-subspace update
using Incremental Partial Least Square (IPLS)~\cite{zeng2014incremental}. 
Hoffman et al.~\cite{hoffman2014continuous} proposed 
another approach for the OUDA problem, 
using Continuous-Manifold-based Adaptation (CMA) 
that formulated the OUDA problem 
as a non-convex optimization problem.
Liu et al.~\cite{liu2020learning} suggested a meta-adaptation framework 
that utilized meta-learning~\cite{finn2017model} to tackle the OUDA problem.
More recently, Kumar et al.~\cite{kumar2020understanding} 
theoretically studied gradual domain adaptation,
where the goal was to adapt the source classifier given unlabeled target data 
that shift gradually in distribution. 
They proved that self-training leverages the gradual domain adaptation 
with small Wasserstein-infinity distance.

Other studies have focused on the applications of the OUDA problem.
Mancini et al.~\cite{mancini2018kitting} adopted 
a Batch Normalization~\cite{ioffe2015batch} (BN) technique 
to tackle the OUDA problem for a robot kitting task.
Wu et al.~\cite{wu2019ace} tackled the OUDA problem 
using memory store and meta-learning on a semantic segmentation task.
Xu et al.~\cite{xu2016hierarchical} suggested an online domain adaptation method 
for Deformable Part-based Model (DPM), which is applicable for Multiple Object Tracking (MOT).

More recently, there have been many studies 
on tackling the test-time adaptation task 
with neural-network-based models.
The test-time adaptation task, 
in addition to the OUDA problem setting, 
assumes that the source data is not accessible during the test time.
The NN-based approaches tackled this task by
updating the model parameters online as unlabeled target data arrives.
Bobu et al.~\cite{bobu2018adapting} proposed a replay method 
to overcome catastrophic forgetting in neural networks.
Wu et al.~\cite{wu2021online} suggested the Online Gradient Descent (OGD) and Follow The History (FTH) techniques for tackling the OUDA problem.
Sun et al.~\cite{sun2020test} proposed the Test-Time-Training (TTT) approach 
that utilizes self-supervised auxiliary task to solve the OUDA problem.
They rotated the test images by 0, 90, 180 and 270 degrees 
and had the model predicted the angle of rotation as a four-way classification problem.
Schneider et al.~\cite{schneider2020improving} removed the covariate shift between the source and the target distribution by batch normalization.
Wang et al.~\cite{DBLP:conf/iclr/WangSLOD21} updated the model by minimizing the entropy loss of the unlabeled target data.

\section{Proposed Approach}
\subsection{Problem Description}
\begin{figure}[htb]
  \centering
  \includegraphics[width=\linewidth]{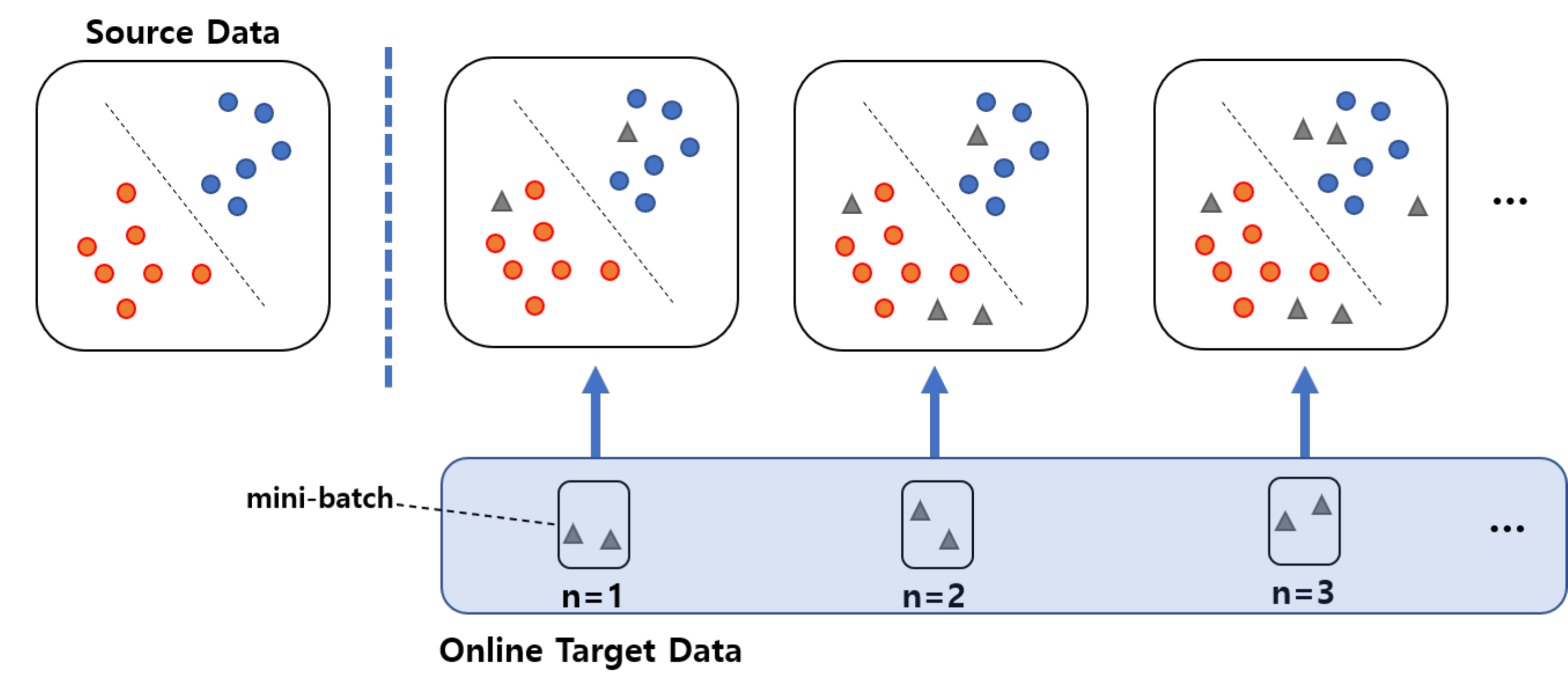}
  \caption{Overview of the OUDA problem. 
  Given the classifier trained 
  with labeled source data (circles) from two classes (orange and blue), 
  the OUDA problem aims
  to classify the unlabeled target data (grey triangles)
  arriving online in a mini-batch.
  Our proposed framework focuses on aligning each target data to the source domain.}
  \label{fig:intro}
\end{figure}

The OUDA problem assumes that the source-domain data are labeled
while the target data are unlabeled 
and arriving online in a sequence of mini-batches at each timestep.
As shown in Fig.~\ref{fig:intro},
the goal of the OUDA problem 
is to classify the arriving unlabeled target mini-batches
with the model pre-trained by the source data.
Formally, the samples in the source domain 
$\mathbf{X}_{\mathcal{S}}\in \mathbb{R}^{N_{\mathcal{S}} \times d}$ 
are given and labeled as 
$\mathbf{Y}_{\mathcal{S}}\in \mathbb{R}^{N_{\mathcal{S}} \times c}$, 
where $N_{\mathcal{S}}$, $d$, and $c$ 
indicate the number of source data, the dimension of the data 
and the number of class categories, respectively.
The target data $\mathbf{X}_{\mathcal{T}}=\{\mathbf{X}_{\mathcal{T},1}, \mathbf{X}_{\mathcal{T},2}, \cdots, \mathbf{X}_{\mathcal{T},B}\}$ 
in the target domain $\mathcal{T}$
are a sequence of unlabeled mini-batches that arrive in an online fashion,
where $B$ indicates the number of mini-batches.
As mentioned previously,
we use the term \textit{mini-batch} for the $n^{th}$ target-data batch
$\mathbf{X}_{\mathcal{T},n}\in \mathbb{R}^{N_{\mathcal{T}} \times d}$,
where $N_{\mathcal{T}}$ indicates the number of data in each mini-batch.
Subscript $(\mathcal{T},n)$ represents 
the $n^{th}$ mini-batch in the target domain.
$N_{\mathcal{T}}$ is assumed to be a constant for $n=1,2,\cdots,B$ 
and is small compared to $N_\mathcal{S}$.
Our goal is to transform 
each target mini-batch $\mathbf{X}_{\mathcal{T},n}$ 
to $\mathbf{X'}_{\mathcal{T},n}$, 
which is aligned with the source domain $\mathcal{S}$, 
in an online fashion. 
The transformed target data 
$\mathbf{X'}_{\mathcal{T},n}$
can be classified correctly as $\mathbf{\hat{Y}}_{\mathcal{T},n}$ 
with the classifier pre-trained in the source domain $\mathcal{S}$. 
The nomenclature of our paper is summarized in Table~\ref{tab:notation}.

\begin{figure*}[!ht]
  \centering
  \includegraphics[width=0.80\linewidth]{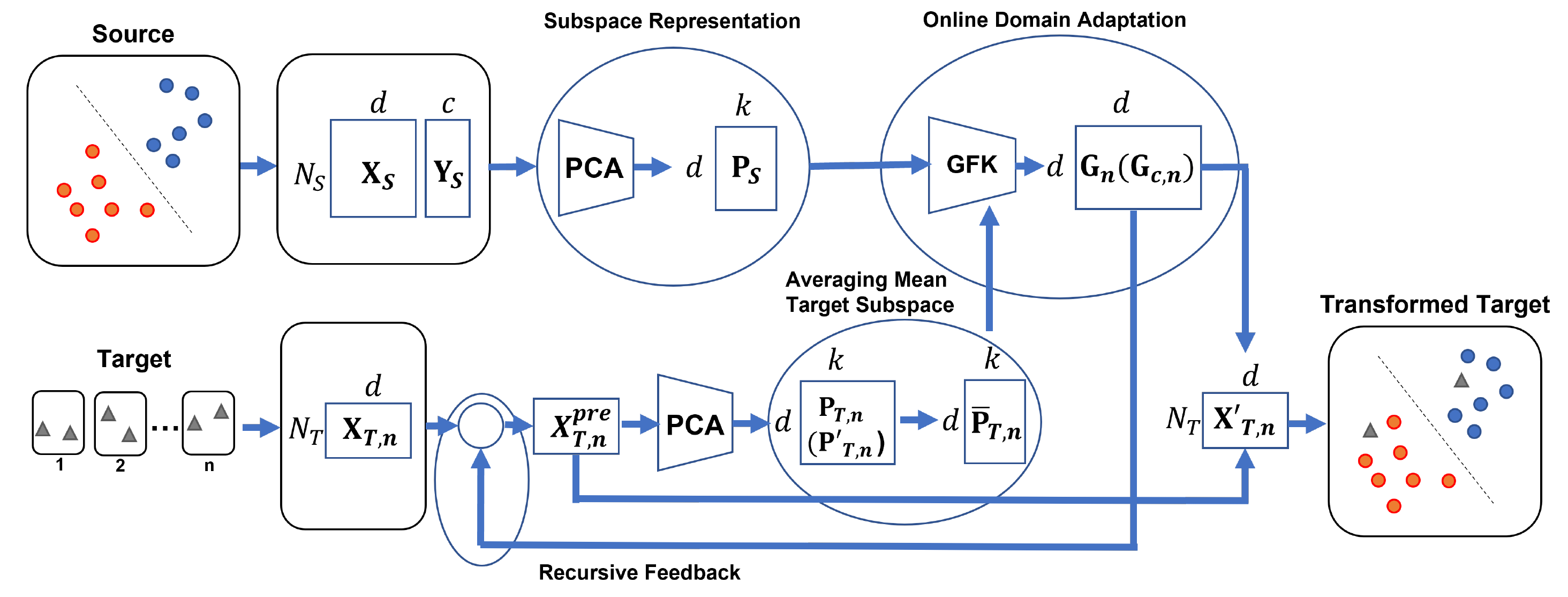}
  \caption{Schematic of the proposed OUDA framework. 
  The proposed framework consists of four stages: 
  1) Subspace representation, 2) Averaging Mean-target subspace, 3) Online domain adaptation, 
  and 4) Recursive feedback.}
\label{fig:schematic}
\vspace*{-0.1in}
\end{figure*}

\begin{table}[!ht]
 \caption{Nomenclature of Proposed Framework}
  \centering
  \resizebox{\columnwidth}{!}{
  \begin{tabular}{l|l}
  \hline
  Notation & Meaning \\
  \hline
  $\mathcal{S}$ & The source domain \\
  $\mathcal{T}$ & The target domain \\
  $N_{\mathcal{S}}$ & The number of the source data samples  \\
  $N_{\mathcal{T}}$ & The number of data samples in a mini-batch \\
  $\mathcal{G}(k, d)$ & $k$-dimensional Grassmann manifold of $\mathbb{R}^{d}$ \\
  $\mathbf{X}_{\mathcal{S}}$ & The set of source data \\
  $\mathbf{Y}_{\mathcal{S}}$ & The source labels \\
  $\mathbf{X}_{\mathcal{T}}$ & The set of target data \\
  $\mathbf{X}_{\mathcal{T},i}$ & $i^{th}$ target mini-batch \\
  $\mathbf{X'}_{\mathcal{T}}$ & The set of transformed target data \\
  $\mathbf{X'}_{\mathcal{T},i}$ & $i^{th}$ transformed target mini-batch \\
  $\mathbf{X}'_{\mathcal{T},i}$ & Transformed $i^{th}$ target mini-batch \\
  $\mathbf{X}^{pre}_{\mathcal{T},n}$ & The pre-processed $n^{th}$ target mini-batch \\
  $\mathbf{P}_{\mathcal{S}}$ & The source subspace of the source data \\
  $\mathbf{P}_{\mathcal{T},i}$ & The target subspace of $i^{th}$ target mini-batch \\
  $\mathbf{P'}_{\mathcal{T},n}$ & Compensated target subspace by subspace prediction \\
  $\mathbf{\overline{P}}_{\mathcal{T},i}$ & Mean-target subspace of $i$ subspaces $\mathbf{P}_{\mathcal{T},1}, \mathbf{P}_{\mathcal{T},2}, \cdots , \mathbf{P}_{\mathcal{T},i}$ \\
  $\mathbf{G}_{n}$ & Transformation matrix that transforms $\mathbf{X}_{\mathcal{T},i}$ \\
  $\mathbf{G}_{c,n}$ & Cumulative transformation matrix that transforms $\mathbf{X}_{\mathcal{T},i}$ \\
  \hline
  \end{tabular}
  }
  \label{tab:notation}
\end{table}

\subsection{Preliminaries: Grassmann Manifold}
Throughout this paper, 
we utilize a Grassmann manifold 
$\mathcal{G}(k, d)$~\cite{edelman1998geometry}~--~
a space that parameterizes 
all $k$-dimensional linear subspaces 
of $d$-dimensional vector space $\mathbb{R}^{d}$.
A single point on $\mathcal{G}(k,d)$ represents a subspace 
that does not depend on the choice of basis. 
Formally, for a $d\! \times\! k $ matrix $\mathbf{P}$ of full rank $k$ 
and any nonsingular $k\! \times\! k$ matrix $\mathbf{L}$, 
the column space col($\mathbf{P}$) of set $\mathbf{A}=\{\mathbf{PL}|\mathbf{P} \in \mathcal{V}(k,d), \mathbf{L} \in SO(k) \}$ 
is a single point on $\mathcal{G}(k,d)$, 
where $\mathcal{V}(k,d)$ is a point on Stiefel manifold~\cite{james1976topology} 
and $SO(k)$ is a special orthogonal group~\cite{edelman1998geometry}.
For simplicity of notation, we denote this point col($\mathbf{P}$) as corresponding basis $\mathbf{P} \in \mathbb{R}^{d \times k}$ throughout this manuscript.

\subsection{Proposed OUDA Framework}
The proposed OUDA framework 
consists of four stages for processing
an incoming $n^{th}$ mini-batch target data:
1) Subspace representation, 
2) Averaging mean-target subspace, 
3) Online domain adaptation, 
and 4) Recursive feedback
as shown in Fig.~\ref{fig:schematic}.
Instead of utilizing the raw samples from both source and target domains,
Stage one embeds those samples 
$\mathbf{X}_{\mathcal{S}}$ and
$\mathbf{X}_{\mathcal{T},n}$
to low-dimensional subspaces 
(i.e. $k$-dimensional subspace of $\mathbb{R}^d$)
$\mathbf{P}_{\mathcal{S}}$ 
and $\mathbf{P}_{\mathcal{T},n}$, respectively, 
for faster computation of transformation matrix
from the target domain $\mathcal{T}$
to the source domain $\mathcal{S}$.

Stage two computes the mean of target subspaces 
$\mathbf{\overline{P}}_{\mathcal{T},n}$ 
embedded in a Grassmann manifold 
using our novel \textit{Incremental Computation of Mean Subspace (ICMS)} method.
The proposed ICMS technique is an efficient method of computing the mean of target subspaces 
and its computed mean is a valid approximate of the Karcher mean~\cite{karcher1977riemannian}.
The $n^{th}$ target subspace $\mathbf{P}_{\mathcal{T},n}$
can be further compensated as $\mathbf{P'}_{\mathcal{T},n}$
by target-subspace prediction
as discussed in a later subsection.
Stage three is the online domain adaptation 
and it computes a transform matrix $\mathbf{G}_{n}$ 
that aligns each arriving target mini-batch to the source domain.
$\mathbf{G}_{n}$ 
can be replaced with the matrix $\mathbf{G}_{c,n}$
that is computed with a cumulative computation technique.
Stage four provides recursive feedback 
by feeding $\mathbf{G}_{n}$ back to the next mini-batch 
$\mathbf{X}_{\mathcal{T},n+1}$.
Each stage is briefly described next.

\subsubsection{Subspace Representation}
The goal of our proposed OUDA framework 
is to find the transformation matrix 
$\mathbf{G}=\{\mathbf{G}_{1}, \mathbf{G}_{2}, \cdots , \mathbf{G}_{B}\}$ 
that transforms the set of target mini-batches 
$\mathbf{X}_{\mathcal{T}}=\{\mathbf{X}_{\mathcal{T},1}, \mathbf{X}_{\mathcal{T},2}, \cdots , \mathbf{X}_{\mathcal{T},B}\}$
to 
$\mathbf{X'}_{\mathcal{T}}=\{\mathbf{X'}_{\mathcal{T},1}, \mathbf{X'}_{\mathcal{T},2}, \cdots , \mathbf{X'}_{\mathcal{T},B}\}$ 
so that the transformed target data $\mathbf{X'}_{\mathcal{T},n}=\mathbf{X}_{\mathcal{T},n}\mathbf{G}_{n}$ ($n=1,2,\cdots,B$)
are well aligned to the source domain, 
where $\mathbf{G}_{n}\in\mathbb{R}^{d \times d}$ indicates 
the transformation matrix from $\mathbf{X}_{\mathcal{T},n}$ 
to $\mathbf{X'}_{\mathcal{T},n}$.
Since all the computations must be conducted online, 
we prefer not to use methods 
that compute $\mathbf{G}_{n}$ directly 
on the original high-dimensional data space. 
Hence, we embed those samples, 
$\mathbf{X}_{\mathcal{S}}$ and
$\mathbf{X}_{\mathcal{T},n}$,
in low-dimensional subspaces
$\mathbf{P}_{\mathcal{S}}=f(\mathbf{X}_{\mathcal{S}})
\in \mathbb{R}^{d \times k}$ 
and $\mathbf{P}_{\mathcal{T},n}=f(\mathbf{X}_{\mathcal{T},n})
\in \mathbb{R}^{d \times k}$, respectively,
where $d$ represents the dimension of the original data 
and $k$ represents the dimension of the subspace.
As mentioned above, we represent the subspace as a corresponding basis matrix.
$f(\cdot)$ can be any low-dimensional representation,
but in this paper we adopt 
the Principal Component Analysis (PCA) algorithm~\cite{wold1987principal}
to obtain $\mathbf{P}_{\mathcal{S}}$ and $\mathbf{P}_{\mathcal{T}}$ 
since it is simple and fast for online domain adaptation 
and is suitable for both labeled and unlabeled data.

\subsubsection{Averaging Mean-target Subspace}
Since a subspace is represented 
as a single point on a Grassmann manifold, 
$\mathbf{P}_{\mathcal{S}}$ and 
$\mathbf{P}_{\mathcal{T},1}, \mathbf{P}_{\mathcal{T},2}, 
\cdots , \mathbf{P}_{\mathcal{T},n}$ 
are represented as $(n+1)$ points on $\mathcal{G}(k, d)$.
Since all the samples 
in the source domain $\mathcal{S}$ are given,
$\mathbf{X}_{\mathcal{S}}$ can be compressed 
to a low-dimensional subspace 
$\mathbf{P}_{\mathcal{S}}$ 
by conducting the embedding technique only once.
For the target domain $\mathcal{T}$, however,
each mini-batch $\mathbf{X}_{\mathcal{T},n}$ should be
represented as a low-dimensional subspace
$\mathbf{P}_{\mathcal{T},n}$
in every timestep as it arrives.
Since a single mini-batch 
is assumed to contain 
a small number of the target samples
(i.e., $N_{\mathcal{T}}$ is small),
it does not sufficiently represent the target domain. 

To find a subspace that generalizes the target domain $\mathcal{T}$,
we compute the mean-target subspace 
$\mathbf{\overline{P}}_{\mathcal{T},n}$ of $n$ target subspaces 
$\mathbf{P}_{\mathcal{T},1}, \mathbf{P}_{\mathcal{T},2}, \cdots, \mathbf{P}_{\mathcal{T},n}$.
Traditionally, Karcher mean~\cite{karcher1977riemannian}
is utilized as the mean of subspaces represented 
on a Grassmann manifold.
Unfortunately, computing the Karcher mean is an iterative process,
which is not suitable for online domain adaptation.
Thus, we  propose a novel computation technique, 
called \textit{Incremental Computation of Mean-Subspace (ICMS)}, 
to compute the mean-target subspace 
$\mathbf{\overline{P}}_{\mathcal{T},i}$ {\em online}
for every $i=1,2, \cdots, n,\cdots, B$.
Different from the iterative process in computing the Karcher mean, 
the proposed ICMS technique incrementally computes 
the mean-target subspace efficiently to satisfy the online computational demand 
and its computed mean is a close approximate to the Karcher mean. 
Formally, when the $n^{th}$ mini-batch $\mathbf{X}_{\mathcal{T}, n}$ 
arrives and is represented as a subspace $\mathbf{P}_{\mathcal{T}, n}$, 
we incrementally compute the mean-target subspace 
$\mathbf{\overline{P}}_{\mathcal{T},n}$ 
using $\mathbf{P}_{\mathcal{T}, n}$ 
and $\mathbf{\overline{P}}_{\mathcal{T},n-1}$, 
where $\mathbf{\overline{P}}_{\mathcal{T},n-1}$ is the mean subspace 
of ($n-1$) target subspaces 
$\mathbf{P}_{\mathcal{T},1}, \mathbf{P}_{\mathcal{T},2}, \cdots , \mathbf{P}_{\mathcal{T},n-1}$.
We mathematically prove that the computed mean from the ICMS computation
is close to the Karcher mean in Section~\ref{section:ICMS}.
$\mathbf{\overline{P}}_{\mathcal{T},n}$ can be rectified to 
$\mathbf{\overline{P}'}_{\mathcal{T},n}$ using the next-target subspace prediction,
which is described in Section~\ref{subsec:NTSP}.

\subsubsection{Online Domain Adaptation}
After the mean-target subspace $\mathbf{\overline{P}}_{\mathcal{T},n}$ 
is computed using the proposed ICMS technique,
we compute the transformation matrix 
$\mathbf{G}_{n}$ that transforms 
the mini-batch $\mathbf{X}_{\mathcal{T},n}$ 
to $\mathbf{X'}_{\mathcal{T},n}$,
where $\mathbf{X'}_{\mathcal{T},n} =\mathbf{X}_{\mathcal{T},n}\mathbf{G}_{n}$.
We adopt the method proposed by Bitarafan et al.~\cite{bitarafan2016incremental}
to compute the matrix $\mathbf{G}_{n}$ 
from $\mathbf{\overline{P}}_{\mathcal{T},n}$
and the source subspace
$\mathbf{P}_{\mathcal{S}}$ using the GFK method~\cite{gong2012geodesic}.
The transformed mini-batch 
$\mathbf{X'}_{\mathcal{T},n}$
is aligned closer to the source domain $\mathcal{S}$ 
compared to the original mini-batch 
$\mathbf{X}_{\mathcal{T},n}$.
This $\mathbf{X'}_{\mathcal{T},n}$ is classified 
by the classifier pre-trained 
with the samples from the source domain $\mathcal{S}$.
Note that $\mathbf{G}_{n}\in \mathbb{R}^{d \times d}$
transforms the original mini-batch
$\mathbf{X}_{\mathcal{T},n}$ and 
not the subspace $\mathbf{P}_{\mathcal{T},n}$.
We exploit this $\mathbf{G}_{n}$ 
when the $(n+1)^{th}$ mini-batch
$\mathbf{X}_{\mathcal{T},n+1}$ arrives.
$\mathbf{G}_{n}$ can be replaced with 
the transformation matrix $\mathbf{G}_{c,n}$, 
which is obtained by cumulative computation 
and is described in Section~\ref{subsec:cumul}.

\subsubsection{Recursive Feedback}\label{subsub:FB}
Followed by the computation of the transformation matrix,
we institute a recursive feedback stage that applies 
the transformation matrix to the next target mini-batch 
before inputting it to the subspace representation stage.
This feedback stage aligns the next target mini-batch 
closer to the source domain before inputting into the subspace representation stage, which leverages the performance of adaptation.
It is crucial to note that the feedback stage at time step $n$
affects the $(n+1)^{th}$ target mini-batch.
Formally, we feed $\mathbf{G}_{n}$ back to 
$\mathbf{X}_{\mathcal{T},n+1}$ as
$\mathbf{X}^{pre}_{\mathcal{T},n+1}
=\mathbf{X}_{\mathcal{T},n+1}\mathbf{G}_{n}$
before inputting $\mathbf{X}_{\mathcal{T},n+1}$ to the first stage of the proposed OUDA framework.
The pre-processed target data $\mathbf{X}^{pre}_{\mathcal{T},n+1}$ is aligned closer to the source subspace $\mathcal{S}$ than $\mathbf{X}_{\mathcal{T},n+1}$.
Combined with next-target subspace prediction, 
the recursive feedback stage leverages the performance of online domain adaptation
when the target domains are noisy.
In the next section, 
we describe the detailed procedure
of the proposed ICMS technique and online domain adaptation.


\section{Incremental Computation of Mean-Subspace for Online Domain Adaptation}
\label{section:ICMS}
The main contribution of the proposed OUDA framework is
the ICMS computation technique followed by online domain adaptation.
The proposed ICMS technique incrementally computes 
the mean-target subspace efficiently on a Grassmann manifold 
and its computed mean is close to the Karcher mean.
The proposed ICMS computation 
also satisfies the online computational demand of the OUDA problem.

\begin{figure}[!ht]
  \centering
  \includegraphics[width=\linewidth]{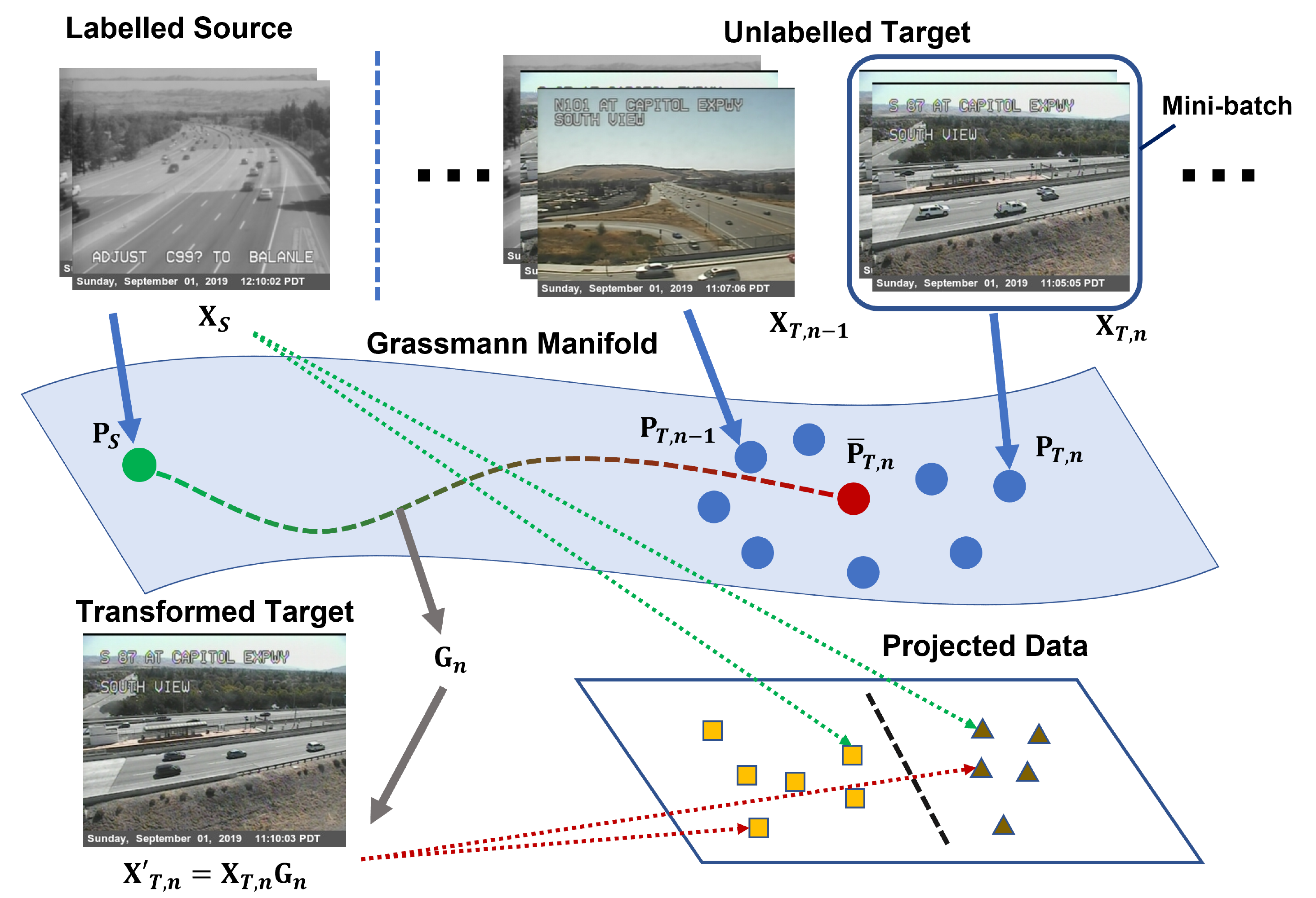}
 \caption{Detailed schematic of the proposed ICMS technique.   
 As the mean-target subspace 
  $\mathbf{\overline{P}}_{\mathcal{T},n}$ (red dot) is computed, 
  the geodesic flow from $\mathbf{P}_{\mathcal{S}}$ (green dot)
  to $\mathbf{\overline{P}}_{\mathcal{T},n}$ is 
  obtained for computing the transformation matrix 
  $\mathbf{G}_{n}$ (grey arrow).
  Transformed target data 
  $\mathbf{X'}_{\mathcal{T},n}$ are then
  projected (red arrow) to a projective space, 
  where the source data $\mathbf{X}_{\mathcal{S}}$ are projected (green arrow).}
\label{fig:detail}
\end{figure}

Figure~\ref{fig:detail} shows the detailed schematic of the proposed ICMS technique.
In Fig.~\ref{fig:detail}, 
as the $n^{th}$ target mini-batch $\mathbf{X}_{\mathcal{T},n}$ 
is represented as the $n^{th}$ target subspace 
$\mathbf{P}_{\mathcal{T},n}$ on a Grassmann manifold,
the mean-target subspace $\mathbf{\overline{P}}_{\mathcal{T},n}$
is efficiently computed on the Grassmann manifold. 
The transformation matrix $\mathbf{G}_{n}$ is computed from the mean-target subspace 
$\mathbf{\overline{P}}_{\mathcal{T},n}$ 
and the source subspace $\mathbf{P}_{\mathcal{S}}$. 
Using this transformation matrix $\mathbf{G}_{n}$, 
the transformed target data $\mathbf{X'}_{\mathcal{T},n}$ is aligned with the source domain.
With this alignment, the classifier trained on the source data 
performs well for the arriving target data.

We first review 
the geodesic flow on a Grassmann manifold,
and then discuss the derivation and computation
of ICMS and its convergence, and finally its utilization for online domain adaptation.

\subsection{Geodesic Flow on a Grassmann Manifold}
A geodesic is the shortest-length curve between two points 
on a manifold~\cite{gallivan2003efficient}. 
The geodesic flow on a Grassmann manifold $\mathcal{G}(k,d)$
that starts from a point $\mathbf{P}_1$ is given 
by $\mathbf{\Phi}(t)=\mathbf{Q}\exp{(t\mathbf{B})}\mathbf{J}$, 
where the matrix $\mathbf{Q}=[\mathbf{P}_{1} \:\:\: \mathbf{R}_{1}] \in SO(d)$,
and $\mathbf{Q}$ is termed as an Orthogonal Completion of $\mathbf{P}_1$ 
such that $\mathbf{Q}^{T}\mathbf{P}_{1}=\mathbf{J}$ and 
$\mathbf{J}=\begin{bmatrix}
\mathbf{I}_{k}\\
\mathbf{O}_{d-k, k}
\end{bmatrix}$. 
Here the matrices $\mathbf{I}_{k}$ and $\mathbf{O}_{d-k, k}$ denote 
$k\! \times\! k$ identity matrix and $(d-k)\! \times\! k$ zero matrix, respectively. 
Using the geodesic parameterization 
with a single parameter $t$~\cite{gallivan2003efficient}, 
the geodesic flow from $\mathbf{P}_{1}$ to $\mathbf{P}_{2}$ on a Grassmann manifold $\mathcal{G}(k,d)$
is parameterized as $\mathbf{\Psi}\!: t\in[0,1]\xrightarrow{}\mathbf{\Psi}(t)\in \mathcal{G}(k,d)$:
\begin{equation}
\label{eq:geodesic}
    \mathbf{\Psi}(t)=\mathbf{P}_{1}\mathbf{U}_{1}\mathbf{\Gamma}(t) - \mathbf{R}_{1}\mathbf{U}_{2}\mathbf{\Sigma}(t)
\end{equation}
under the two constraints that $\mathbf{\Psi}(0)=\mathbf{P}_{1}$ and $\mathbf{\Psi}(1)=\mathbf{P}_{2}$, 
where $\mathbf{\Gamma}(t)$ and 
$\mathbf{\Sigma}(t)=[\mathbf{\Sigma}_{1}(t)^{T} \text{     } \mathbf{O}^{T}]^{T}$ 
are diagonal and block diagonal matrices 
whose elements are $\cos{(t\theta_{i,n})}$ 
and $\sin{(t\theta_{i,n})}$, respectively.
$\mathbf{R}_{1}$ denotes the orthogonal complement to $\mathbf{P}_{1}$, 
namely $\mathbf{R}_{1}^{T}\mathbf{P}_{1}=\mathbf{O}$.
Two orthonormal matrices 
$\mathbf{U}_{1}
\in \mathbb{R}^{k \times k}$ and 
$\mathbf{U}_{2}
\in \mathbb{R}^{(d-k) \times (d-k)}$ 
are given by the following Generalized Singular-Value Decompositions (GSVD)~\cite{paige1981towards},
\begin{align}
\label{eq:SVD_GFK_general}
    \mathbf{Q}^{T}\mathbf{P}_{2}=
    \begin{pmatrix}
    \mathbf{U}_{1} & \mathbf{0} \\
    \mathbf{0} & \mathbf{U}_{2}
    \end{pmatrix}
    \begin{pmatrix}
    \mathbf{\Gamma}(t)\\
    \mathbf{-\Sigma}(t)
    \end{pmatrix}
    \mathbf{V}^{T} ,
\end{align}
where $\mathbf{\Gamma}(t) \in \mathbb{R}^{k \times k}$ and $\mathbf{\Sigma}(t)=[\mathbf{\Sigma}_{1}(t)^{T} \text{     } \mathbf{O}^{T}]^{T} \in \mathbb{R}^{(d-k) \times k}$ are diagonal and block diagonal matrices, respectively, 
and $\mathbf{\Sigma}_{1}(t) \in \mathbb{R}^{k \times k}$ 
and $\mathbf{O} \in \mathbb{R}^{(d-2k) \times k}$.

\subsection{Mean-target Subspace Computation -- ICMS}
\label{subsection:mean}
Since computing the Karcher mean is 
a time-consuming, iterative process and not suitable for 
online domain adaptation, 
we propose a novel technique, 
called \textit{Incremental Computation of Mean-Subspace (ICMS)},
for computing the mean-target subspace
of $n$ target subspaces on a Grassmann manifold.

\begin{figure}[!t]
  \centering
  \subfloat[]{
  \includegraphics[width=0.41\linewidth]{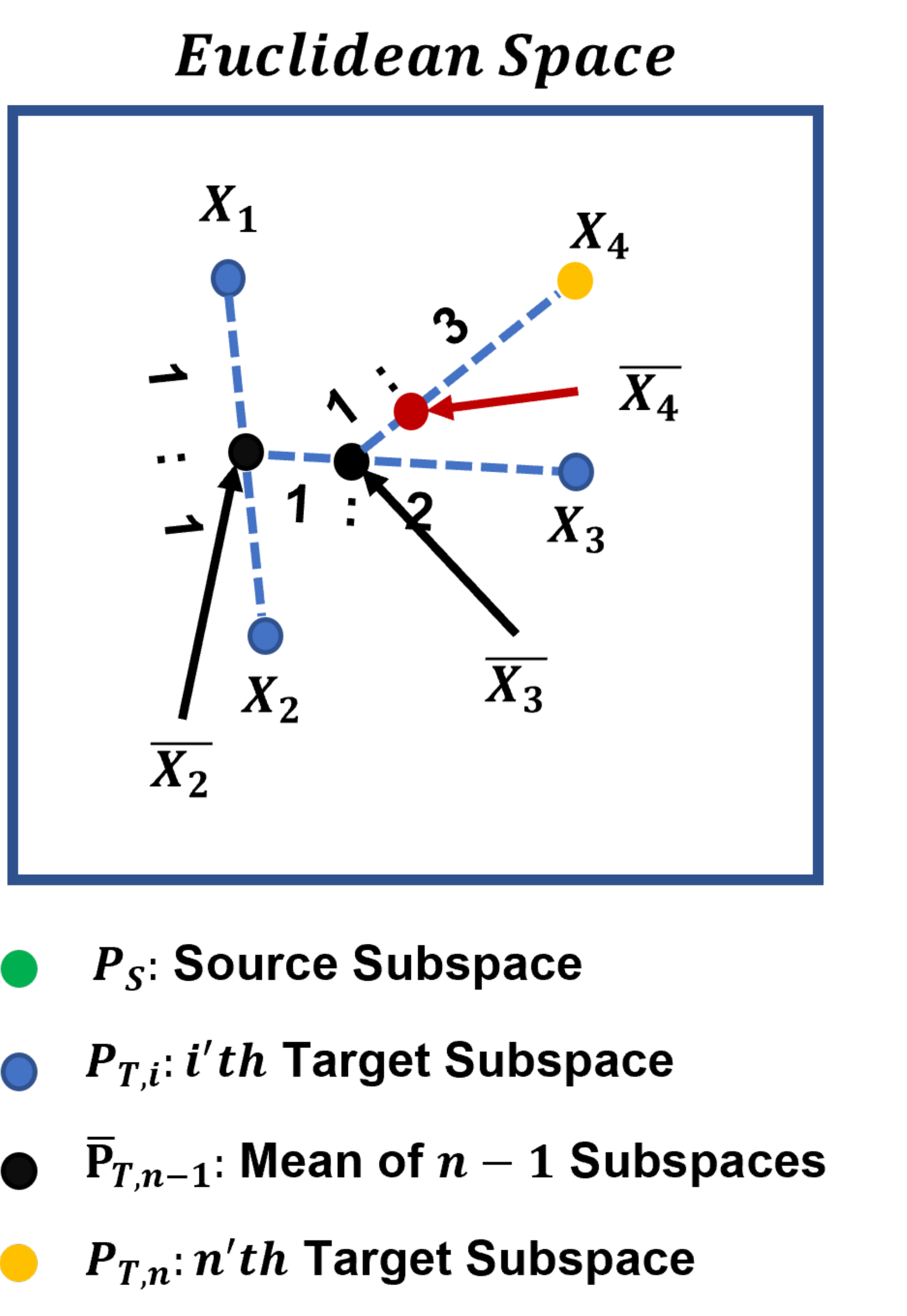}}
  \hfill
  \subfloat[
  ]
 {\includegraphics[width=0.57\linewidth]{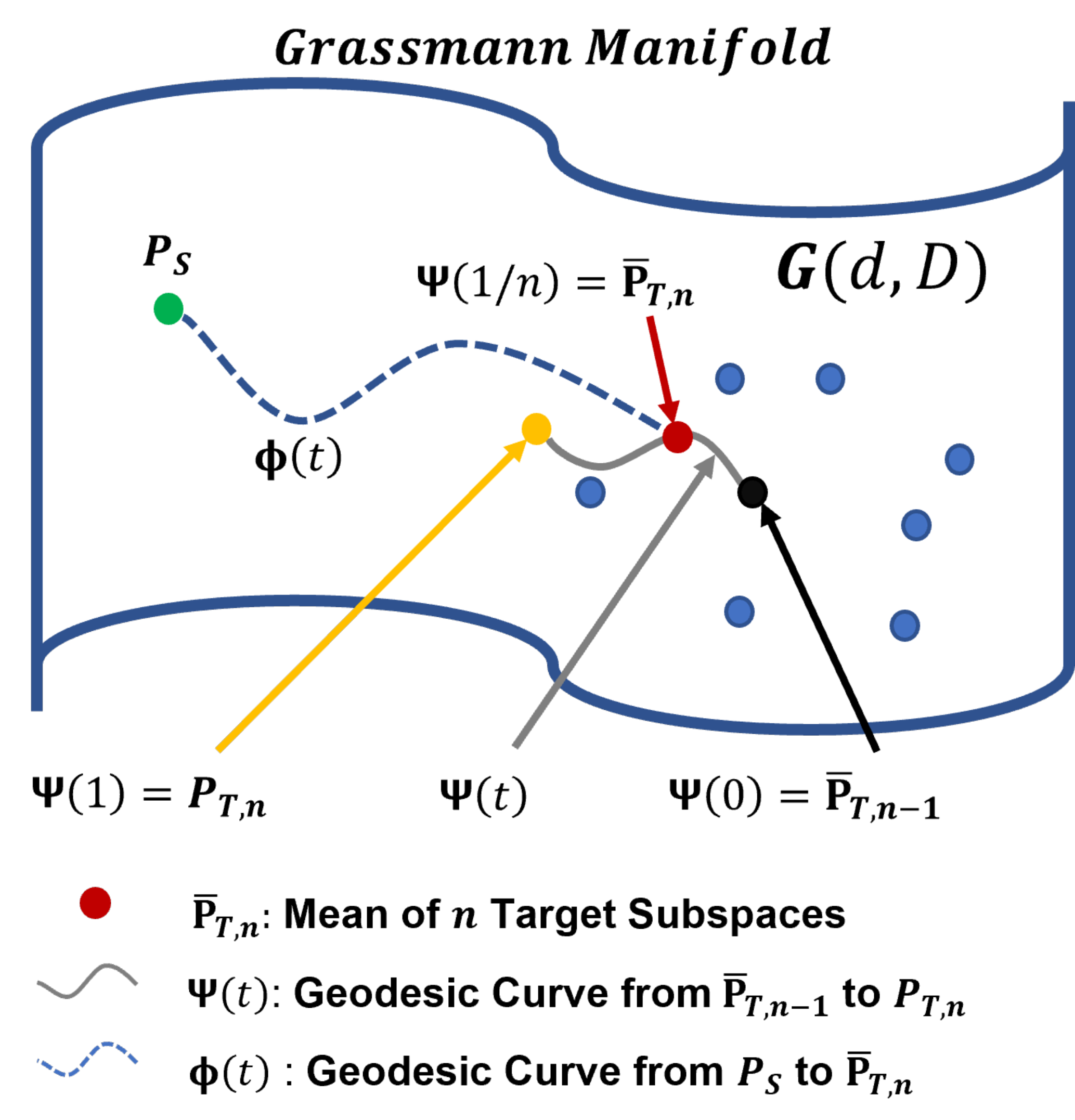}}
 \caption{Mean-target subspace computation.~~ 
 (a) Geometric interpretation of computing the mean point of $n$ points in the Euclidean space.~~
 (b) Computation of mean-target subspace 
  $\mathbf{\overline{P}}_{\mathcal{T},n}$ on a Grassmann manifold.}
\label{fig:geometric}
\vspace*{-0.1in}
\end{figure}

The proposed ICMS is inspired by the geometric interpretation 
of computing the mean of $n$ points on the Euclidean space.
As shown in Fig.~\ref{fig:geometric}(a), 
it is geometrically intuitive 
to compute the mean point $\mathbf{\overline{X}}_{n}$ 
of $n$ points $\mathbf{X}_{1}, \mathbf{X}_{2}, \cdots , \mathbf{X}_{n}$ 
in an incremental way when the points are on the Euclidean space.
If the mean point $\mathbf{\overline{X}}_{n-1}$ of  ($n-1$) points 
$\mathbf{X}_{1}, \mathbf{X}_{2}, \cdots , \mathbf{X}_{n-1}$ 
as well as the $n^{th}$ point $\mathbf{X}_{n}$ are given, 
the updated mean point $\mathbf{\overline{X}}_{n}$ can be computed as 
$\mathbf{\overline{X}}_{n}=\{(n-1)\mathbf{\overline{X}}_{n-1}+\mathbf{X}_{n}\}/n$.
From a geometric perspective, 
$\mathbf{\overline{X}}_{n}$ is the internal point, 
where the distances from 
$\mathbf{\overline{X}}_{n}$ to $\mathbf{\overline{X}}_{n-1}$ 
and to $\mathbf{X}_{n}$ have the ratio of $1\!:\!(n-1)$:
\begin{equation}
\label{eq:Euclidean}
    |\mathbf{\overline{X}}_{n-1}\mathbf{\overline{X}}_{n}|
    =\frac{|\mathbf{\overline{X}}_{n-1}\mathbf{X}_{n}|}{n}.    
\end{equation}

We adopt this geometric perspective of ratio concept 
to the Grassmann manifold. 
However, Eq.~\eqref{eq:Euclidean} is not directly applicable
to a Grassmann manifold
since the distance between two points on the Grassmann manifold is not an Euclidean distance.
In Fig.~\ref{fig:geometric}(b), 
we update the mean-target subspace $\mathbf{\overline{P}}_{\mathcal{T},n}$ 
of $n$ target subspaces when the previous mean 
subspace $\mathbf{\overline{P}}_{\mathcal{T},n-1}$ of ($n-1$) target subspaces 
and the $n^{th}$ subspace $\mathbf{P}_{\mathcal{T},n}$ are given.
$\mathbf{\overline{P}}_{\mathcal{T},n}$ is the introspection point 
that divides the geodesic flow 
from $\mathbf{\overline{P}}_{\mathcal{T},n-1}$ 
to $\mathbf{P}_{\mathcal{T},n}$ to the ratio of $1\!:\!(n-1)$.
Following Eq.~(\ref{eq:geodesic}), 
the geodesic flow from 
$\mathbf{\overline{P}}_{\mathcal{T},n-1}$ 
to $\mathbf{P}_{\mathcal{T},n}$ 
is parameterized as 
$\mathbf{\Psi}_{n}:t\in[0,1]\xrightarrow{}\mathbf{\Psi}_{n}(t)\in G(k,d)$:
\begin{equation}
\label{eq:ICMS}
    \mathbf{\Psi}_{n}(t)=\mathbf{\overline{P}}_{\mathcal{T},n-1}\mathbf{U}_{1,n}\mathbf{\Gamma}_{n}(t) - \mathbf{\overline{R}}_{\mathcal{T},n-1}\mathbf{U}_{2,n}\mathbf{\Sigma}_{n}(t)
\end{equation}
under the constraints of  $\mathbf{\Psi}_{n}(0)\!=\!\mathbf{\overline{P}}_{\mathcal{T},n-1}$  
and $\mathbf{\Psi}_{n}(1)\!=\!\mathbf{P}_{\mathcal{T},n}$. 
$\mathbf{\overline{R}}_{\mathcal{T},n-1}\in\mathbb{R}^{d\times(d-k)}$ 
denotes the orthogonal complement to 
$\mathbf{\overline{P}}_{\mathcal{T},n-1}$; 
that is, $\mathbf{\overline{R}}_{\mathcal{T},n-1}^{T}\mathbf{\overline{P}}_{\mathcal{T},n-1}=\mathbf{0}$. 
Two orthonormal matrices $\mathbf{U}_{1,n}\in \mathbb{R}^{k \times k}$ 
and $\mathbf{U}_{2,n}\in \mathbb{R}^{(d-k) \times (d-k)}$ 
are defined similar to Eq.~(\ref{eq:SVD_GFK_general}) and
given by the following pair of singular-value decompositions (SVDs),
\begin{align}
\label{eq:SVD_mean}
    \mathbf{\overline{P}}_{\mathcal{T},n-1}^{T}\mathbf{P}_{\mathcal{T},n}=\mathbf{U}_{1,n}\mathbf{\Gamma}_{n}\mathbf{V}_{n}^{T} \:\: \\ 
    \mathbf{\overline{R}}_{\mathcal{T},n-1}^{T}\mathbf{P}_{\mathcal{T},n}=-\mathbf{U}_{2,n}\mathbf{\Sigma}_{n}\mathbf{V}_{n}^{T}
\end{align}
where $\mathbf{\Gamma}_{n} \in \mathbb{R}^{k \times k}$ and $\mathbf{\Sigma}_{n}=[\mathbf{\Sigma}_{1,n}^{T} \text{     } \mathbf{O}^{T}]^{T} \in \mathbb{R}^{(d-k) \times k}$ are diagonal and block diagonal matrices, respectively, 
and $\mathbf{\Sigma}_{1,n} \in \mathbb{R}^{k \times k}$ and $\mathbf{O} \in \mathbb{R}^{(d-2k) \times k}$.

When the $n^{th}$ target mini-batch $\mathbf{X}_{\mathcal{T}, n}$ 
arrives and is represented as the subspace $\mathbf{P}_{\mathcal{T}, n}$, 
we incrementally compute the mean-target subspace 
$\mathbf{\overline{P}}_{\mathcal{T},n}$ 
using $\mathbf{P}_{\mathcal{T}, n}$ 
and $\mathbf{\overline{P}}_{\mathcal{T},n-1}$, 
where $\mathbf{\overline{P}}_{\mathcal{T},n-1}$ is the mean subspace 
of ($n-1$) target subspaces 
$\mathbf{P}_{\mathcal{T},1}, \mathbf{P}_{\mathcal{T},2}, \cdots , \mathbf{P}_{\mathcal{T},n-1}$.
Finally, utilizing $\mathbf{\Psi}_{n}(t)$ in Eq.~\eqref{eq:ICMS}, 
we obtain $\mathbf{\overline{P}}_{\mathcal{T},n}=\mathbf{\Psi}_{n}(\frac{1}{n})$ and 
the mean-target subspace of $n$ target subspaces 
$\mathbf{\overline{P}}_{\mathcal{T},n}=\mathbf{\Psi}_{n}(\frac{1}{n})$ 
can be incrementally computed as:
\begin{equation}
\label{eq:mean_target_subspace}
    \mathbf{\overline{P}}_{\mathcal{T},n}=\mathbf{\overline{P}}_{\mathcal{T},n-1}\mathbf{U}_{1,n}\mathbf{\Gamma}_{n}(\frac{1}{n}) - \mathbf{\overline{R}}_{\mathcal{T},n-1}\mathbf{U}_{2,n}\mathbf{\Sigma}_{n}(\frac{1}{n})   .
\end{equation}
Note that $n$ refers to the $n^{th}$ mini-batch in the target domain. 
Since $0 \leq \frac{1}{n} \leq 1$,  
$\mathbf{\Gamma}_{n}(\frac{1}{n})$ 
and $\mathbf{\Sigma}_{n}(\frac{1}{n})$ 
are well defined.

\subsection{Convergence of Mean-target Subspace}
In this subsection,
we prove that the ICMS-based computed mean in Eq.~\eqref{eq:mean_target_subspace}
on a Grassmann manifold 
is a valid approximation to the Karcher mean.
The proof is conducted by induction on $n$.

We assume that the mean-subspace 
$\mathbf{\overline{P}}_{\mathcal{T},n-1}$ of $(n-1)$ subspaces 
$\mathbf{P}_{\mathcal{T}, 1}, \mathbf{P}_{\mathcal{T},2}, \cdots, \mathbf{P}_{\mathcal{T}, n-1}$
on a Grassmann manifold $\mathcal{G}$
is close to the Karcher mean.
We then want to show that $\mathbf{\overline{P}}_{\mathcal{T},n}$, 
the computed mean subspace of $n$ subspaces $\mathbf{P}_{\mathcal{T},1}$, 
$\mathbf{P}_{\mathcal{T},2}, \cdots , \mathbf{P}_{\mathcal{T},n}$, 
is also close to the Karcher mean by showing that:
\begin{equation}
\label{eq:wts}
    \sum_{i=1}^{n} \overrightarrow{\mathbf{\overline{P}}_{\mathcal{T},n}\mathbf{P}_{\mathcal{T},i}} \simeq \mathbf{0} .
\end{equation}
Satisfying Eq.~\eqref{eq:wts} is sufficient to prove that the computed mean subspace 
$\mathbf{\overline{P}}_{\mathcal{T},n}$ is close to the Karcher mean.
Since the $n$ subspaces $\mathbf{P}_{\mathcal{T}, 1}$,
$\mathbf{P}_{\mathcal{T}, 2}, \cdots , \mathbf{P}_{\mathcal{T}, n}$ and the computed mean subspace 
$\mathbf{\overline{P}}_{\mathcal{T},n}$ are assumed to be close, 
these subspaces approximately follow the geometrical property in the Euclidean space.

For a large $n > N$, the tangent spaces of the manifold $\mathcal{G}$ 
at the two points $\mathbf{\overline{P}}_{\mathcal{T},n-1}$ and 
$\mathbf{\overline{P}}_{\mathcal{T},n}$ are similar. 
Therefore, Eq.~(\ref{eq:wts}) can be rewritten as follow:
\begin{equation}
\label{eq:wts2}
\begin{split}
    \sum_{i=1}^{n} \overrightarrow{\mathbf{\overline{P}}_{\mathcal{T},n}\mathbf{P}_{\mathcal{T},i}} & 
    = \sum_{i=1}^{n-1} \overrightarrow{\mathbf{\overline{P}}_{\mathcal{T},n}\mathbf{P}_{\mathcal{T},i}} + 
    \overrightarrow{\mathbf{\overline{P}}_{\mathcal{T},n}\mathbf{P}_{\mathcal{T}, n}} \\
    & = \sum_{i=1}^{n} (\overrightarrow{\mathbf{\overline{P}}_{\mathcal{T},n}\mathbf{\overline{P}}_{\mathcal{T},n-1}} + \overrightarrow{\mathbf{\overline{P}}_{\mathcal{T},n-1}\mathbf{P}_{\mathcal{T}, i}}) \\
    & \quad + \overrightarrow{\mathbf{\overline{P}}_{\mathcal{T},n}\mathbf{P}_{\mathcal{T}, n}} \\
    & \simeq (n-1)\overrightarrow{\mathbf{\overline{P}}_{\mathcal{T},n}\mathbf{\overline{P}}_{\mathcal{T},n-1}} + 
    \overrightarrow{\mathbf{\overline{P}}_{\mathcal{T},n}\mathbf{P}_{\mathcal{T}, n}}  .
\end{split}
\end{equation}
Let the geodesic flow from $\mathbf{\overline{P}}_{\mathcal{T},n-1}$ 
to $\mathbf{P}_{\mathcal{T}, n}$ be given as
$\mathbf{\Psi}_{n}(t)=\mathbf{\overline{Q}}_{n-1}\exp{(t\mathbf{B})}\mathbf{J}$. 
The matrix $\mathbf{\overline{Q}}_{n-1} \in SO(d)$ such that 
$\mathbf{\overline{Q}}_{n-1}^{T}\overline{\mathbf{P}}_{\mathcal{T},n-1}=\mathbf{J}$ 
and $\mathbf{J}=
\begin{bmatrix}
\mathbf{I}_{k}\\
\mathbf{O}_{d-k, k}
\end{bmatrix}$, 
where $\mathbf{B}$ is a skew-symmetric matrix.
The mean subspace $\mathbf{\overline{P}}_{\mathcal{T},n}$ 
is a point on the geodesic as 
$\mathbf{\overline{P}}_{\mathcal{T},n}=\mathbf{\Psi}_{n}(\frac{1}{n})$.
To prove Eq.~(\ref{eq:wts2}),
we re-parameterize the two geodesics $\mathbf{S}_{1}(t)$ 
and $\mathbf{S}_{2}(t)$ 
starting from $\mathbf{\overline{P}}_{\mathcal{T},n}$ to $\overline{\mathbf{P}}_{\mathcal{T},n-1}$ and $\mathbf{P}_{\mathcal{T},n}$, respectively:

\begin{equation}
    \begin{split}
        \mathbf{S}_{1}(t) & = \mathbf{\Psi}_{n}(\frac{1-t}{n}) = \mathbf{\overline{Q}}_{n-1}\exp{(\frac{1-t}{n}\mathbf{B})}\mathbf{J} \\
        & = \mathbf{\overline{Q}}_{n-1}\exp{(\frac{\mathbf{B}}{n})}
        \exp{(t(-\frac{\mathbf{B}}{n}))}\mathbf{J}  
    \end{split}
\end{equation}

\begin{equation}
    \begin{split}
        \mathbf{S}_{2}(t) & = \mathbf{\Psi}_{n}(\frac{(n-1)t+1}{n}) = 
        \mathbf{\overline{Q}}_{n-1}\exp{(\frac{(n-1)t+1}{n}\mathbf{B})}\mathbf{J} \\
        & = \mathbf{\overline{Q}}_{n-1}\exp{(\frac{\mathbf{B}}{n})}
        \exp{(t(-\frac{n-1}{n})\mathbf{B})}\mathbf{J}.
    \end{split}
\end{equation}
From these two equations, we can derive that\\
$\dot{\mathbf{S}}_{1}(t) = -\frac{1}{n}\dot{\mathbf{\Psi}}_{n}(\frac{1-t}{n})$ and
$\dot{\mathbf{S}}_{2}(t) = \frac{n-1}{n}\dot{\mathbf{\Psi}}_{n}(\frac{(n-1)t+1}{n})$.\\
Hence, the first and third terms in Eq.~(\ref{eq:wts2}) can be rewritten as:
\begin{equation}
\label{eq:pnpn-1}
    \overrightarrow{\mathbf{\overline{P}}_{\mathcal{T},n}\mathbf{\overline{P}}_{\mathcal{T},n-1}} = \dot{\mathbf{S}}_{1}(0) = -\frac{1}{n}\dot{\mathbf{\Psi}}_{n}(\frac{1}{n})  
\end{equation}
\begin{equation}
\label{eq:pnpn}
    \overrightarrow{\mathbf{\overline{P}}_{\mathcal{T},n}\mathbf{P}_{\mathcal{T}, n}} = \dot{\mathbf{S}}_{2}(0) = \frac{n-1}{n}\dot{\mathbf{\Psi}}_{n}(\frac{1}{n})  .
\end{equation}
Substituting Eqs.~(\ref{eq:pnpn-1}) and (\ref{eq:pnpn}) 
into Eq.~(\ref{eq:wts2}), we obtain
\begin{equation}
    (n-1)\overrightarrow{\mathbf{\overline{P}}_{\mathcal{T},n}\mathbf{\overline{P}}_{\mathcal{T},n-1}} + 
\overrightarrow{\mathbf{\overline{P}}_{\mathcal{T},n}\mathbf{P}_{\mathcal{T}, n}} = \mathbf{0} . 
\end{equation}
Hence,
$$\sum_{i=1}^{n} \overrightarrow{\mathbf{\overline{P}}_{\mathcal{T},n}\mathbf{P}_{\mathcal{T},i}} \simeq \mathbf{0} ~~ \text {for} ~~{\forall}{n}>N \text{     }_{\square}$$
\noindent
Thus, the mean subspace computed by ICMS is close to the Karcher mean.

\subsection{Next-Target Subspace Prediction}\label{subsec:NTSP}
\begin{figure}[!ht]
  \centering
  \includegraphics[width=0.6\linewidth]{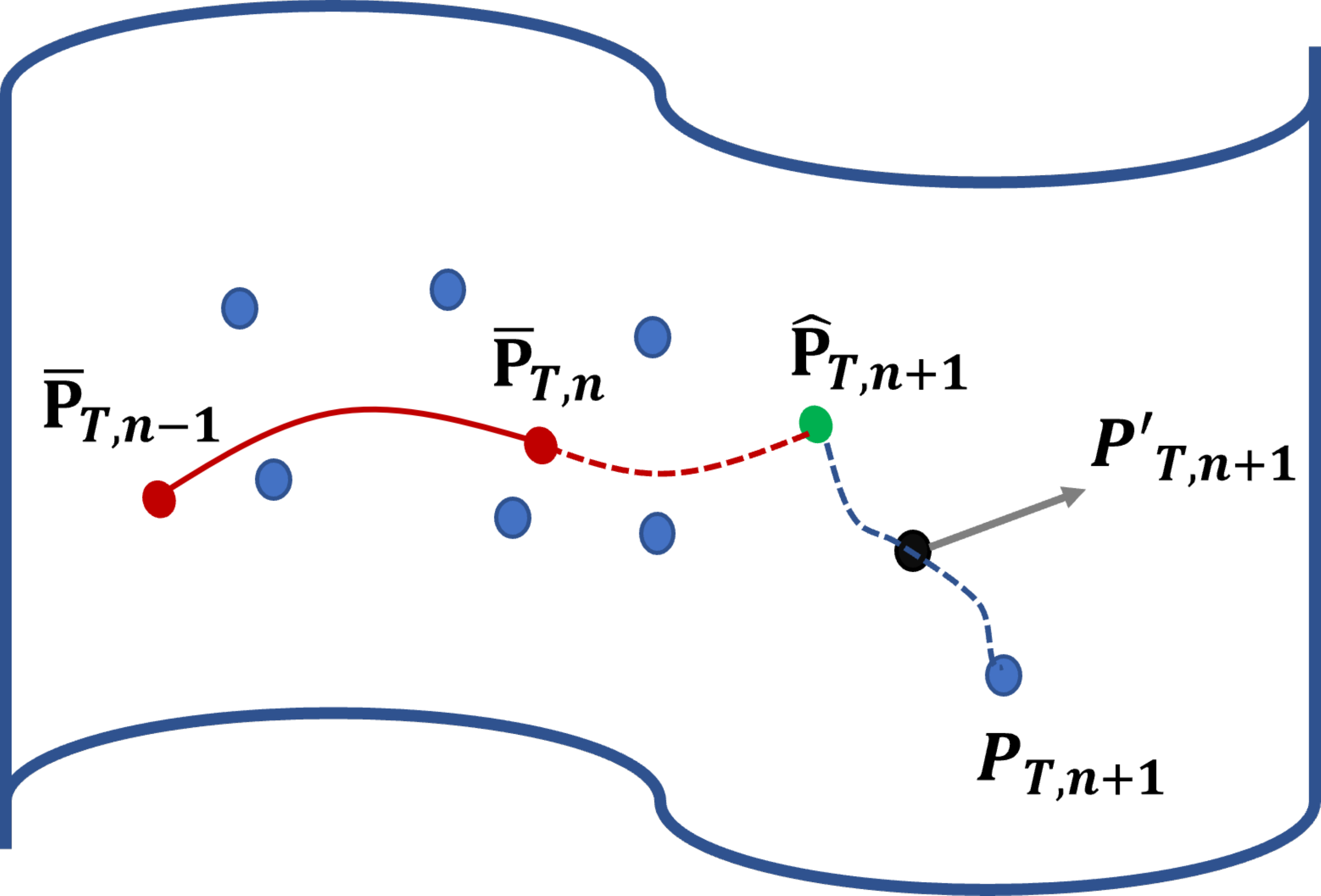}
 \caption{
 Prediction of next-target subspace.
 The $(n+1)^{th}$ target subspace 
 $\mathbf{\hat{P}}_{\mathcal{T},n+1}$ (green dot) 
 is an extrapolation of the geodesic 
 from the $(n-1)^{th}$ mean-target subspace $\mathbf{\overline{P}}_{\mathcal{T},n-1}$ 
 to the $n^{th}$ mean-target subspace $\mathbf{\overline{P}}_{\mathcal{T},n}$.
 The $(n+1)^{th}$ target subspace $\mathbf{P}_{\mathcal{T},n+1}$ 
 can then be compensated to $\mathbf{P'}_{\mathcal{T},n+1}$ (black dot).}
\label{fig:pred}
\end{figure}

The target subspace $\mathbf{P}_{\mathcal{T}, n}$ 
is computed at every timestep as the $n^{th}$ mini-batch 
$\mathbf{X}_{\mathcal{T}, n}$ arrives.
However, $\mathbf{X}_{\mathcal{T}, n}$ may be noisy 
due to many factors such as imbalanced data distribution 
and data contamination,
leading to an inaccurate target subspace $\mathbf{P}_{\mathcal{T}, n}$
and the mean-target subspace $\mathbf{\overline{P}}_{\mathcal{T},n}$.
To rectify this inaccurate target subspace,
we predict the next-target subspace from the flow of the mean target subspaces.
Formally, we propose to compute the prediction of next-target subspace $\mathbf{\hat{P}}_{\mathcal{T},n+1}$
by using the previous and current mean-target subspaces 
$\mathbf{\overline{P}}_{\mathcal{T},n-1}$ and
$\mathbf{\overline{P}}_{\mathcal{T},n}$, respectively, 
as shown in Fig.~\ref{fig:pred}. 
For the OUDA problem with continuously changing environment,
we assume that the mean-target subspace may shift 
according to a continuous curve on the Grassmann manifold.
We compute the velocity matrix $\mathbf{A}_{n}$ 
of the geodesic from the previous mean-target subspace 
$\mathbf{\overline{P}}_{\mathcal{T},n-1}$
to the current mean-target subspace $\mathbf{\overline{P}}_{\mathcal{T},n}$.
We then obtain the prediction of the next-target subspace 
$\mathbf{\hat{P}}_{\mathcal{T},n+1}$ by extrapolating the curve
from the obtained velocity matrix $\mathbf{A}_{n}$.
The velocity matrix is computed by utilizing the technique proposed by Gopalan et al.~\cite{gopalan2011domain}.
We first compute the orthogonal completion $\mathbf{Q}_{n-1}$, $\mathbf{Q}_{n}$ of
the mean-target subspaces $\mathbf{\overline{P}}_{\mathcal{T},n-1}$, $\mathbf{\overline{P}}_{\mathcal{T},n}$, respectively.
As in Eq.~(\ref{eq:SVD_GFK_general}),
GSVD of $\mathbf{Q}_{n-1}^{T}\mathbf{\overline{P}}_{\mathcal{T},n}$ is computed as:
\begin{align}
\label{eq:GSVD_QPbar}
    \mathbf{Q}_{n-1}^{T}\mathbf{\overline{P}}_{\mathcal{T},n}=
    \begin{pmatrix}
        \mathbf{U'}_{1, n} & \mathbf{0} \\
        \mathbf{0} & \mathbf{U'}_{2, n}
    \end{pmatrix}
    \begin{pmatrix}
        \mathbf{\Gamma'}_{n}\\
        \mathbf{-\Sigma'}_{n}
    \end{pmatrix}
    \mathbf{V'}^{T}_{n}.
\end{align}

The principal angle between $\mathbf{\overline{P}}_{\mathcal{T},n-1}$
and $\mathbf{\overline{P}}_{\mathcal{T},n}$ 
is computed from $\mathbf{\Theta}_{n}\!=\!\arccos{(\mathbf{\Gamma'}_{n})}$.
Hence, the velocity matrix of the geodesic from $\mathbf{\overline{P}}_{\mathcal{T},n-1}$
to $\mathbf{\overline{P}}_{\mathcal{T},n}$ is as follows:
\begin{align}
    \mathbf{A}_{n}=\mathbf{U'}_{2,n}\mathbf{\Theta}_{n}\mathbf{U'}_{1,n}^{T}.
\end{align}
The predicted next-target subspace $\mathbf{\hat{P}}_{\mathcal{T},n+1}$ of 
$(n+1)^{th}$ mini-batch is the subspace obtained by 
the extrapolation of the geodesic from the
$n^{th}$ mean-target subspace $\mathbf{\overline{P}}_{\mathcal{T},n}$
with velocity matrix $\mathbf{A}_{n}$:
\begin{align}
\label{eq:Phat_T}
    \mathbf{\hat{P}}_{\mathcal{T},n+1}=\mathbf{Q}_{n}
    \begin{pmatrix}
        \mathbf{U'}_{1,n}\mathbf{\Gamma'}_{n} \\
        -\mathbf{U'}_{2,n}\mathbf{\Sigma'}_{n}
    \end{pmatrix}.
\end{align}
\noindent
The procedure of computing the next-target subspace 
$\mathbf{\hat{P}}_{\mathcal{T},n+1}$
is described in Algorithm~\ref{alg:next}.

\begin{algorithm}[!ht]
\SetAlgoLined
 \textbf{Input:} 
 \\
 \quad $\mathbf{\overline{P}}_{\mathcal{T},n-1}$: Previous mean-target subspace
 \\
 \quad $\mathbf{\overline{P}}_{\mathcal{T},n}$:~~~ Current mean-target subspace
 \\
 \textbf{Output:} 
 \\
 \quad $\mathbf{\hat{P}}_{\mathcal{T},n+1}$: Prediction of next-target subspace
 \\
 \quad
 \\
 \textbf{Procedure:} 
\\
\quad 1) Compute the orthogonal completion $\mathbf{Q}_{n-1}$, $\mathbf{Q}_{n}$ of
\\
\quad \quad $\mathbf{\overline{P}}_{\mathcal{T},n-1}$, $\mathbf{\overline{P}}_{\mathcal{T},n}$, respectively.
\\
\quad 2) Compute the CS decomposition of $\mathbf{Q}_{n-1}^{T}\mathbf{\overline{P}}_{\mathcal{T},n}$ 
\\
\quad\quad by Eq.~(\ref{eq:GSVD_QPbar}).
\\
\quad 3) Compute the principal angle matrix: \\
\quad\quad $\mathbf{\Theta}_{n}=\arccos{(\mathbf{\Gamma'}_{n})}$~.
\\
\quad 4) Compute the velocity matrix: \\
\quad\quad $\mathbf{A}_{n}=\mathbf{U'}_{2,n}\mathbf{\Theta}_{n}\mathbf{U'}_{1,n}^{T}$~.
\\
\quad 5) Predict the next-target subspace by Eq.~(\ref{eq:Phat_T})~.
\caption{Prediction of Next-Target Subspace}
\label{alg:next}
\end{algorithm}

The predicted next-target subspace $\mathbf{\hat{P}}_{\mathcal{T},n+1}$ 
is compensated to $\mathbf{P'}_{\mathcal{T},n+1}$ 
as the actual target subspace of $(n+1)^{th}$ mini-batch 
$\mathbf{P}_{\mathcal{T},n+1}$ arrives.
The compensated target subspace $\mathbf{P'}_{\mathcal{T}, n+1}$ 
is the introspection of the prediction $\mathbf{\hat{P}}_{\mathcal{T},n+1}$ 
and the observation $\mathbf{P}_{\mathcal{T}, n+1}$.

\subsection{Online Domain Adaptation}
\label{section:DA}
In this stage, 
we compute the transformation matrix 
$\mathbf{G}_{n}$ 
using the GFK method~\cite{gong2012geodesic}. 
The transformation matrix $\mathbf{G}_{n}$ is computed from 
the source subspace $\mathbf{P}_{\mathcal{S}}$ 
and the $n^{th}$ mean-target subspace 
$\mathbf{\overline{P}}_{\mathcal{T},n}$;
that is, 
$\mathbf{G}_{n}=GFK(\mathbf{P}_{\mathcal{S}}, \mathbf{\overline{P}}_{\mathcal{T},n})$.
After computing the mean-target subspace 
$\mathbf{\overline{P}}_{\mathcal{T},n}$, 
we parameterize a geodesic flow from $\mathbf{P}_{\mathcal{S}}$ 
to $\mathbf{\overline{P}}_{\mathcal{T},n}$ 
as $\mathbf{\Phi}_{n}:t\in[0,1]\xrightarrow{}
\mathbf{\Phi}_{n}(t)\in \mathcal{G}(k,d)$:

\begin{equation}
\label{eq:GFK}
    \mathbf{\Phi}_{n}(t)=\mathbf{P}_{\mathcal{S}}\mathbf{U}_{3,n}\mathbf{\Lambda}_{n}(t) - \mathbf{R}_{\mathcal{S}}\mathbf{U}_{4,n}\mathbf{\Omega}_{n}(t)
\end{equation}
under the constraints of $\mathbf{\Phi}_{n}(0)=\mathbf{P}_{\mathcal{S}}$ 
and $\mathbf{\Phi}_{n}(1)=\mathbf{\overline{P}}_{\mathcal{T},n}$. $\mathbf{R}_{\mathcal{S}}\in\mathbb{R}^{d\times(d-k)}$ 
denotes the orthogonal complement 
to $\mathbf{P}_{\mathcal{S}}$; that is,  $\mathbf{R}_{\mathcal{S}}^{T}\mathbf{P}_{\mathcal{S}}=\mathbf{0}$.
Two orthonormal matrices 
$\mathbf{U}_{3,n}\in \mathbb{R}^{k \times k}$ 
and $\mathbf{U}_{4,n}\in \mathbb{R}^{(d-k) \times (d-k)}$ 
are given by the GSVD\cite{paige1981towards},
\begin{align}
\label{eq:SVD_GFK}
    \mathbf{Q}_{\mathcal{S}}^{T}\mathbf{\overline{P}}_{\mathcal{T},n}=
    \begin{pmatrix}
    \mathbf{U}_{3,n} & \mathbf{0} \\
    \mathbf{0} & \mathbf{U}_{4,n}
    \end{pmatrix}
    \begin{pmatrix}
    \mathbf{\Lambda}_{n}(t)\\
    \mathbf{-\Omega}_{n}(t)
    \end{pmatrix}
    \mathbf{W}_{n}^{T}~.
\end{align}
Based on the GFK, 
the transformation matrix $\mathbf{G}_{n}$ 
from the target domain 
to the source domain 
is found by projecting and integrating 
over the infinite set 
of all intermediate subspaces between them:
\begin{equation}
\label{eq:GFK_proj}
    \int_{0}^{1}(\mathbf{\Phi}_{n}(\alpha)^{T}\mathbf{x}_{i})^{T}(\mathbf{\Phi}_{n}(\alpha)^{T}\mathbf{x}_{j})d\alpha=\mathbf{x}_{i}^{T}\mathbf{G}_{n}\mathbf{x}_{j} .
\end{equation}
From the above equation, 
we can derive the closed form of $\mathbf{G}_{n}$ as:
\begin{equation}
\label{eq:GFK_DA}
    \mathbf{G}_{n}=\int_{0}^{1}\mathbf{\Phi}_{n}(\alpha)\mathbf{\Phi}_{n}(\alpha)^{T}d\alpha .
\end{equation}

We adopt this $\mathbf{G}_{n}$ 
as the transformation matrix 
to the preprocessed target data as $\mathbf{X'}_{\mathcal{T},n}=\mathbf{X}^{pre}_{\mathcal{T},n}\mathbf{G}_{n}$, 
which better aligns the target data to the source domain.
$\mathbf{X}^{pre}_{\mathcal{T},n}$ is the target data 
fed back from the previous mini-batch.

\subsection{Cumulative Computation of Transformation Matrix, $\mathbf{G}_{c,n}$}
\label{subsec:cumul}
\begin{figure}[!t]
  \centering
  \includegraphics[width=0.8\linewidth]{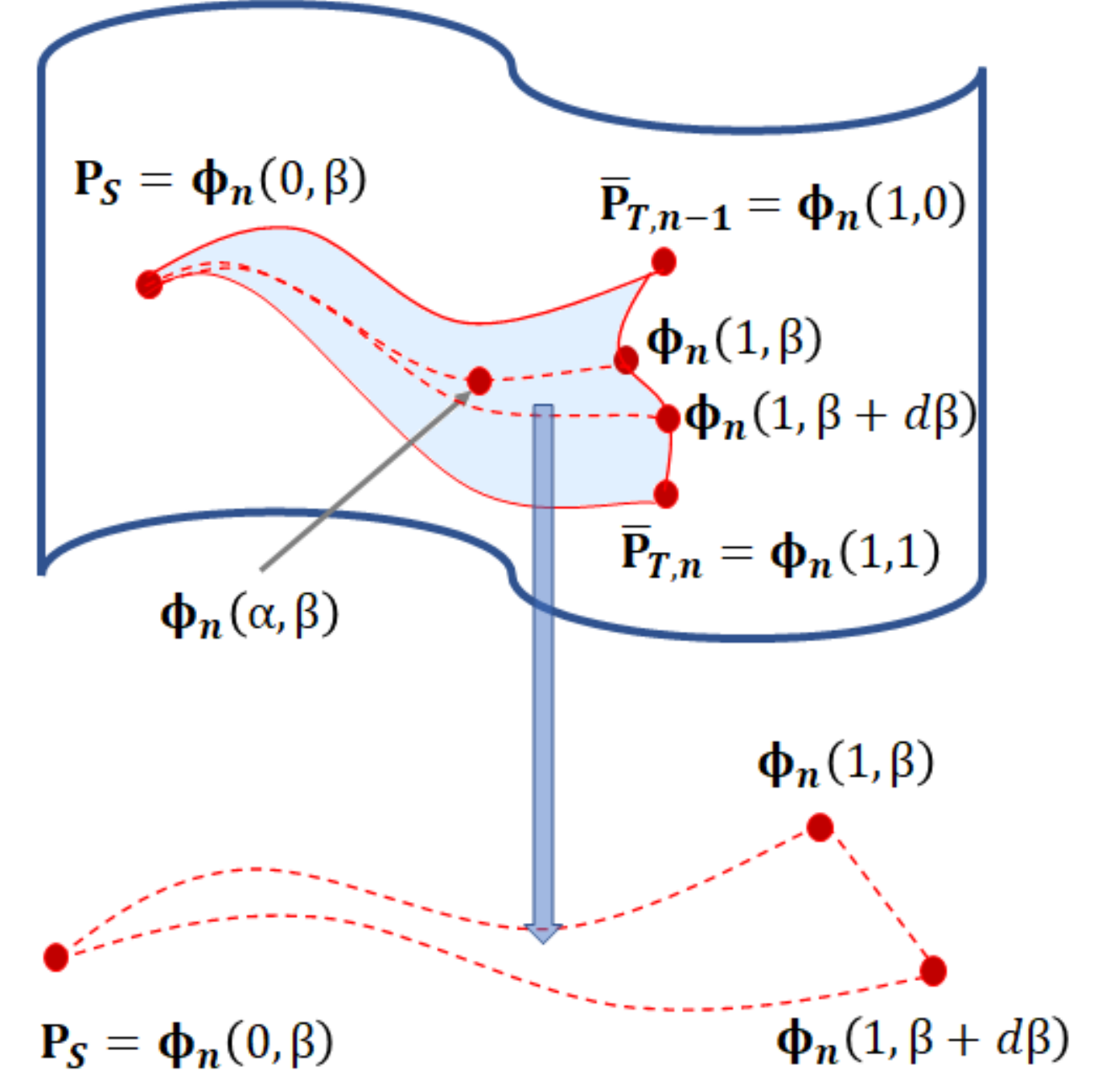}
 \caption{Cumulative computation of transformation matrix 
 $\mathbf{G}_{c,n}$.
 $\mathbf{G}_{c,n}$ is obtained by integrating over the area bounded by 
 the source subspace $\mathbf{P}_{\mathcal{S}}$, 
 $(n-1)^{th}$ mean-target subspace $\mathbf{\overline{P}}_{\mathcal{T},n-1}$ and 
 $n^{th}$ mean-target subspace $\mathbf{\overline{P}}_{\mathcal{T},n}$ (enclosed and colored area).
 The subspace inside the area is parameterized as $\mathbf{\Phi}_{n}(\alpha, \beta)$.}
\label{fig:cumul}
\vspace*{-0.1in}
\end{figure}

The proposed ICMS method updates the mean-target subspace 
as each  target mini-batch arrives, 
but the transformation matrix $\mathbf{G}_{n}$ 
is still computed by merely utilizing the source subspace 
$\mathbf{P}_{\mathcal{S}}$ and the mean-target subspace 
$\mathbf{\overline{P}}_{\mathcal{T},n}$.
To obtain the transformation matrix 
that embraces the cumulative temporal dependency,
we propose a method of computing the cumulative transformation matrix 
$\mathbf{G}_{c,n}$.
As shown in Fig.~\ref{fig:cumul}, 
we compute the cumulative transformation matrix $\mathbf{G}_{c,n}$ 
by considering the variation of the mean-target subspaces caused by two consecutive mini-batches.
The cumulative transformation matrix $\mathbf{G}_{c,n}$ is computed based on the area 
bounded by the three points $\mathbf{P}_{\mathcal{S}}$, 
$\mathbf{\overline{P}}_{\mathcal{T},n-1}$, and $\mathbf{\overline{P}}_{\mathcal{T},n}$ on the manifold.
In the OUDA problem, it is assumed that 
the target domain is shifted continuously but slowly.
Hence, the principal angle $\mathbf{\Theta}_{n}(0)$ between the source subspace 
$\mathbf{P}_{\mathcal{S}}$ and the previous mean-target subspace 
$\mathbf{\overline{P}}_{\mathcal{T},n-1}$ changes linearly 
to the principal angle $\mathbf{\Theta}_{n}(1)$ between the source subspace $\mathbf{P}_{\mathcal{S}}$ 
and the current mean-target subspace $\mathbf{\overline{P}}_{\mathcal{T},n}$.
The principal angle between $\mathbf{P}_{\mathcal{S}}$ 
and the intermediate subspace is denoted as follows:
\begin{align}
    \mathbf{\Theta}_{n}(\beta) = \mathbf{\Theta}_{n}(0) + 
    (\mathbf{\Theta}_{n}(1) - \mathbf{\Theta}_{n}(0))\beta .
\end{align}
To consider all the intermediate subspaces inside the boundary area,
double integration is conducted for the computation instead of using Eq.~(\ref{eq:GFK_proj}):
\begin{equation}
    \int_{0}^{1}\int_{0}^{1}(\mathbf{\Phi}_{n}(\alpha, \beta)^{T}\mathbf{x}_{i})^{T}(\mathbf{\Phi}_{n}(\alpha, \beta)^{T}\mathbf{x}_{j})d\alpha d\beta=\mathbf{x}_{i}^{T}\mathbf{G}_{c, n}\mathbf{x}_{j} .
\end{equation}
Hence, $\mathbf{G}_{c,n}$ is computed as:
\begin{equation}
\label{eq:Gcn_int}
    \mathbf{G}_{c, n}=\int_{0}^{1}\int_{0}^{1}\mathbf{\Phi}_{n}(\alpha, \beta)\mathbf{\Phi}_{n}(\alpha, \beta)^{T}d\alpha d\beta
    = \int_{0}^{1}\mathbf{G}_{n}(\beta)d\beta.
\end{equation}

Different from the offline domain-adaptation problem~\cite{gong2012geodesic}, 
the transformation matrix $\mathbf{G}_{n}$ 
depends on the parameter $\beta$:
\begin{equation}
\label{eq:Gn_beta}
    \mathbf{G}_{n}(\beta)=[\mathbf{P}_{\mathcal{S}}\mathbf{U}_{3,n} \:\:
    \mathbf{R}_{\mathcal{S}}\mathbf{U}_{4,n}]
    \begin{bmatrix}
        \mathbf{\Lambda}_{1,n}(\beta) & \mathbf{\Lambda}_{2,n}(\beta) \\
        \mathbf{\Lambda}_{2,n}(\beta) & \mathbf{\Lambda}_{3,n}(\beta) 
    \end{bmatrix}
    \begin{bmatrix}
        \mathbf{U}_{3,n}^{T}\mathbf{P}_{\mathcal{S}}^{T} \\
        \mathbf{U}_{4,n}^{T}\mathbf{R}_{\mathcal{S}}^{T} 
    \end{bmatrix} .
\end{equation}
The computed $\mathbf{G}_{n}(\beta)$ is then integrated 
over the parameter $\beta$, as shown in the bottom part of Fig.~\ref{fig:cumul}.
Utilizing the GFK technique~\cite{gong2012geodesic}, 
the diagonal elements of matrices $\mathbf{\Lambda}_{1,n}(\beta)$, $\mathbf{\Lambda}_{2,n}(\beta)$, and $\mathbf{\Lambda}_{3,n}(\beta)$ 
are, respectively:
\begin{align}
\label{eq:lambda}
    \nonumber
    \lambda_{1i,n}(\beta)&=1 + \frac{\sin{2\theta_{i,n}(\beta)}}{2\theta_{i,n}(\beta)}, \\ 
    \nonumber
    \lambda_{2i,n}(\beta)&=\frac{\cos{2\theta_{i,n}(\beta) - 1}}{2\theta_{i,n}(\beta)}, \\
    \lambda_{3i,n}(\beta)&=1 - \frac{\sin{2\theta_{i,n}(\beta)}}{2\theta_{i,n}(\beta)} .
\end{align}

Substituting Eq.~(\ref{eq:Gn_beta}) into Eq.~(\ref{eq:Gcn_int}),
only the second matrix of the right-hand side of Eq.~(\ref{eq:Gn_beta}) 
should be integrated with respect to $\beta$. Thus,
\begin{equation}
\label{eq:Gcn_cumul}
    \mathbf{G}_{c,n}=[\mathbf{P}_{\mathcal{S}}\mathbf{U}_{3,n} \:\:
    \mathbf{R}_{\mathcal{S}}\mathbf{U}_{4,n}]
    \begin{bmatrix}
        \mathbf{\Delta}_{1,n} & \mathbf{\Delta}_{2,n} \\
        \mathbf{\Delta}_{2,n} & \mathbf{\Delta}_{3,n} 
    \end{bmatrix}
    \begin{bmatrix}
        \mathbf{U}_{3,n}^{T}\mathbf{P}_{\mathcal{S}}^{T} \\
        \mathbf{U}_{4,n}^{T}\mathbf{R}_{\mathcal{S}}^{T} 
    \end{bmatrix} ,
\end{equation}
where the diagonal elements of  matrices $\mathbf{\Delta}_{1}$, $\mathbf{\Delta}_{2}$, 
and $\mathbf{\Delta}_{3}$ are, respectively:
\begin{align}\label{eq:delta}
    \nonumber
    \delta_{1i,n}&=2-\frac{2}{9}\Big\{\theta_{i,n}(1)^{2} + \theta_{i,n}(1)\theta_{i,n}(0) + \theta_{i,n}(0)^{2}\Big\}, \\
    \nonumber
    \delta_{2i,n}&=-\frac{1}{2}\Big\{ \theta_{i,n}(1) + \theta_{i,n}(0) \Big\} , 
     \\
    \delta_{3i,n}&=\frac{2}{9}\Big\{\theta_{i,n}(1)^{2} + \theta_{i,n}(1)\theta_{i,n}(0) + \theta_{i,n}(0)^{2}\Big\} .
\end{align}
The computation of the cumulative transformation matrix $\mathbf{G}_{c,n}$
is summarized in Algorithm~\ref{alg:cumul}.

\begin{algorithm}[!ht]
\SetAlgoLined
 \textbf{Input:} 
 \\
 \quad $\mathbf{P}_{\mathcal{S}}$: Source subspace
 \\
 \quad $\mathbf{\overline{P}}_{\mathcal{T},n-1}$: Previous mean-target subspace
 \\
 \quad $\mathbf{\overline{P}}_{\mathcal{T},n}$: Current mean-target subspace
 \\
 \textbf{Output:} 
 \\
 \quad $\mathbf{G}_{c,n}$: Prediction of next-target subspace
 \\
 \quad
 \\
 \textbf{Procedure:} 
\\
\quad 1) Compute the orthogonal completion $\mathbf{Q}_{\mathcal{S}}$, $\mathbf{Q}_{n-1}$, 
\\
\quad \quad $\mathbf{Q}_{n}$ of $\mathbf{P}_{\mathcal{S}}$, $\mathbf{\overline{P}}_{\mathcal{T},n-1}$, $\mathbf{\overline{P}}_{\mathcal{T},n}$, respectively.
\\
\quad 2) Compute the principal angle matrices $\mathbf{\Theta}_{n}(0)$,\\
\quad\quad $\mathbf{\Theta}_{n}(1)$.
\\
\quad 3) Compute the cumulative matrix $\mathbf{G}_{c,n}$ by Eq.~(\ref{eq:Gcn_cumul}).
\caption{Cumulative Transformation Matrix $\mathbf{G}_{c,n}$ Computation}
\label{alg:cumul}
\end{algorithm}

\subsection{Overall Procedure}
\begin{figure}[htb]
  \centering
  \includegraphics[width=\linewidth]{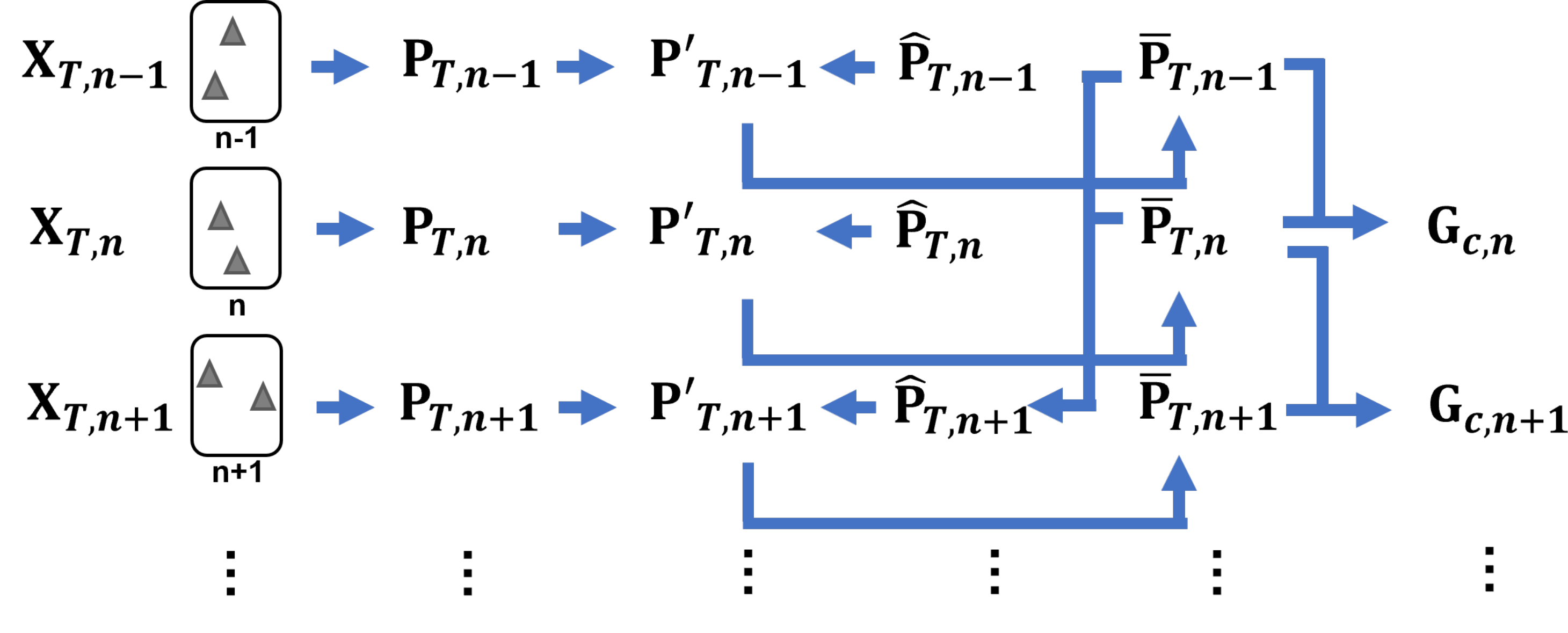}
  \caption{Schematic of subspace prediction and cumulative computation of transformation matrix $\mathbf{G}_{c,n}$.}
\label{fig:pred_cumul_schematic}
\vspace*{0.1in}
\end{figure}

The overall procedure of our proposed multi-stage OUDA framework 
is outlined in Algorithm~\ref{alg:OUDA}.
As shown in Fig.~\ref{fig:pred_cumul_schematic}, 
the prediction of the $i^{th}$ target subspace 
$\mathbf{\hat{P}}_{\mathcal{T}, i}$ 
is computed from the
$(i-2)^{th}$ and $(i-1)^{th}$ mean-target subspaces 
$\mathbf{\overline{P}}_{\mathcal{T},i-2}$ 
and $\mathbf{\overline{P}}_{\mathcal{T},i-1}$, respectively,
whereas the $i^{th}$ cumulative transformation matrix $\mathbf{G}_{c,i}$
is computed from the $(i-1)^{th}$ and $i^{th}$ mean-target subspaces 
$\mathbf{\overline{P}}_{\mathcal{T},i-1}$ and $\mathbf{\overline{P}}_{\mathcal{T},i}$, respectively.

\begin{algorithm}[ht]
\SetAlgoLined
 \textbf{Input:} 
 \\
 \quad $\mathbf{X}_{\mathcal{S}}$: Source data
 \\
 \quad $\mathbf{Y}_{\mathcal{S}}$: Source labels
 \\
 \quad $\mathbf{X}_{\mathcal{T},n}$: $n^{th}$ target data mini-batch
 \\
 \quad $\mathbf{\overline{P}}_{\mathcal{T},n-1}$: Previous mean-target subspace
 \\
\textbf{Output:} 
 \\
\quad $\mathbf{\hat{Y}}_{\mathcal{T},n}$: Predicted label of $\mathbf{X}_{\mathcal{T},n}$
\\
\quad
 \\
\textbf{Procedure:} 
 \\
 \quad 1) Compute the subspaces $\mathbf{P}_{\mathcal{S}}$, $\mathbf{P}_{\mathcal{T},n}$, $\mathbf{P}_{\mathcal{T},n-1}$ from \\
\quad\quad $\mathbf{X}_{\mathcal{S}}$, $\mathbf{X}_{\mathcal{T},n}$, $\mathbf{X}_{\mathcal{T},n-1}$, respectively.
 \\
 \quad 2) Predict the target subspace $\mathbf{\hat{P}}_{\mathcal{T},n}$ using 
 \\
 \quad\quad Algorithm~\ref{alg:next}. \\
 \quad 3) Compensate the next-target subspace $\mathbf{P'}_{\mathcal{T},n}$ 
 \\
 \quad\quad from $\mathbf{\hat{P}}_{\mathcal{T},n}$ and $\mathbf{P}_{\mathcal{T},n}$.
 \\
 \quad 4) Compute the mean-target subspace $\mathbf{\overline{P}}_{\mathcal{T},n}$ by 
 \\
 \quad\quad Eq.~(\ref{eq:mean_target_subspace}).
 \\
 \quad 5) Compute $\mathbf{G}_{c, n}$ from $\mathbf{P}_{\mathcal{S}}$, $\mathbf{\overline{P}}_{\mathcal{T},n-1}$, and $\mathbf{\overline{P}}_{\mathcal{T},n}$.
 \\
 \quad 6) Compute $\mathbf{X'}_{\mathcal{T},n}=\mathbf{X}_{\mathcal{T},n}\mathbf{G}_{c, n}$.
 \\
 \quad 7) Predict the label $\mathbf{\hat{Y}}_{\mathcal{T},n}$ by the classifier trained 
 \\
 \quad\quad with $\mathbf{X}_{\mathcal{S}}$ and $\mathbf{Y}_{\mathcal{S}}$.
 \\
 \quad 8) Update the classifier with $\mathbf{X}_{\mathcal{T},n}$ and $\mathbf{\hat{Y}}_{\mathcal{T},n}$.
 
\caption{Proposed Multi-stage OUDA Framework}
\label{alg:OUDA}
\end{algorithm}

\section{Experimental Results}
We performed extensive computer simulations 
on five small-scale datasets~\cite{bitarafan2016incremental} and one large-scale dataset~\cite{hendrycks2019benchmarking}
to validate the performance of our proposed multi-stage OUDA framework 
in the context of data classification. 
We first evaluated the major components of the proposed OUDA framework, 
namely the ICMS technique and the recursive feedback 
with cumulative computation of the transformation matrix, 
$\mathbf{G}_{c,n}$,
and the next-target subspace prediction.
We then verified the effect of adaptive classifier in our proposed framework.
We also analyzed how the various factors such as
the number of source data $N_{\mathcal{S}}$, 
the target mini-batch $N_{\mathcal{T}}$, 
and the dimension of projected subspace $k$
would affect the classification performance.
To show the validity of our proposed framework, 
we illustrated the convergence of our proposed ICMS technique 
and visualized the projected features.
We then validated the performance of our proposed OUDA framework 
by comparing it to other manifold-based traditional methods
in the context of data classification.
We selected two existing manifold-based traditional methods for comparisons 
~-- ~Evolving Domain Adaptation (EDA)~\cite{bitarafan2016incremental} 
and Continuous Manifold Alignment (CMA)~\cite{hoffman2014continuous}.
The CMA method has two variations depending on the domain adaptation techniques~--~
GFK~\cite{gong2012geodesic} and Statistical Alignment (SA)~\cite{fernando2013unsupervised}.

Furthermore, we also conducted experiments 
on test-time adaptation tasks~\cite{DBLP:conf/iclr/WangSLOD21},
comparing the performance of our proposed framework 
with recent Neural-Network-(NN)-based learning models.

\subsection{Datasets}
\begin{figure}[hbt]
  \centering
  \subfloat[\label{fig:dataset_car}]{
    \includegraphics[width=0.90\linewidth]{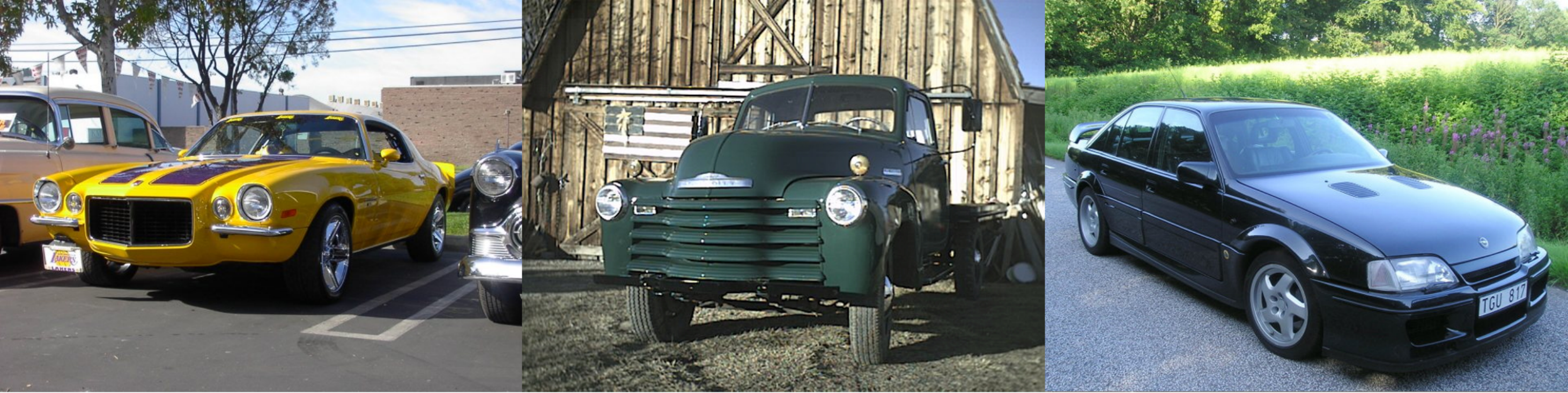}}
    \\
  \subfloat[\label{fig:dataset_traffic}]{
    \includegraphics[width=0.90\linewidth]{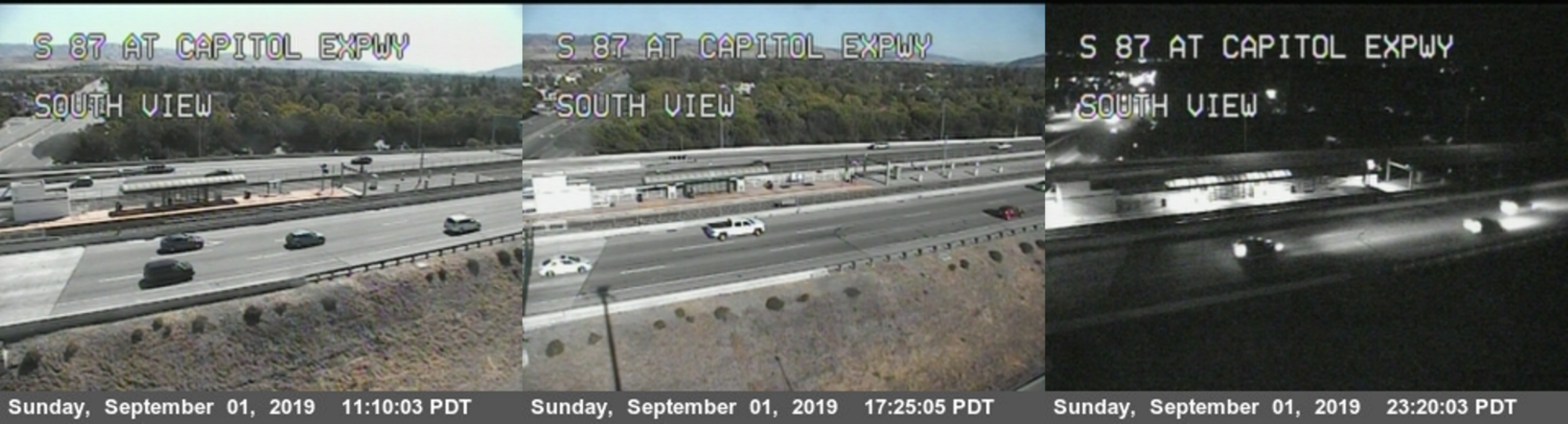}}
  \caption{Image samples of the (a) Car and (b) Traffic datasets.}
  \label{fig:dataset}
\end{figure}

The six datasets~\cite{bitarafan2016incremental} 
that we have selected are -- 
the Traffic dataset, 
the Car dataset,
the Waveform21 dataset, 
the Waveform40 dataset, 
the Weather dataset, 
and the CIFAR-10-C dataset. 
These datasets provide a large variety of time-variant images 
and signals to test upon. 
The Traffic dataset includes images captured from 
a fixed traffic camera observing a road over a 2-week period. 
This dataset consists of 5412 instances 
of 512-dimensional ($d\!=\!512$) GIST features~\cite{oliva2001modeling} 
with two classes as either ``heavy traffic" or ``light traffic."
The Car dataset contains images of automobiles 
manufactured between 1950 and 1999 acquired from online databases. 
This Car dataset includes 1770 instances 
of 4096-dimensional ($d\!=\!4096$) DeCAF~\cite{donahue2014decaf} features 
with two classes as ``sedans" or ``trucks."  
Figure~\ref{fig:dataset}(a) depicts example images of sedans or trucks from 1950's to 1990's. 
Figure~\ref{fig:dataset}(b) depicts the same road 
but the scene changes as the environment changes 
from the morning (left) to afternoon (middle) and night (right).
The Waveform21 dataset is composed of 5000 wave instances 
of 21-dimensional features with three classes.   
The Waveform40 dataset is the second version of the Waveform21 
with additional features. 
This dataset consists of 40-dimensional features.
The Weather dataset includes 18159 daily readings of attributes 
such as temperature, pressure and wind speed.
Those attributes are represented as 8-dimensional features with two classes 
as ``rain" or ``no rain." 
The CIFAR-10-C dataset is an extension of CIFAR-10 dataset that includes 10 categories of images of 50,000 train images and 10,000 test images. The CIFAR-10 dataset is turned into CIFAR-10-C dataset with 15 types of corruptions at 5 severity levels~\cite{hendrycks2019benchmarking}.
The information of these six datasets is summarized in Table~\ref{tab:datasets}. 
Dimension of the subspace was assigned as 
$k=100$ for the datasets of which $d$ is greater than 200, 
or $k=\frac{d}{2}$ otherwise. 

\begin{table}[htb]
 \caption{The table of datasets}
  \centering
  \resizebox{\columnwidth}{!}{
  \begin{tabular}{ccccccc}
  \hline
  Method & Traffic & Car & Wave21 & Wave40 & Weather & CIFAR-10-C \\
  \hline
  Number of samples & 5412 & 1770 & 5000 & 5000 & 18159 & 150000 \\
  Dimension of data($d$) & 512 & 4096 & 21 & 40 & 8 & - \\
  Default $k$ & 100 & 100 & 10 & 20 & 4 & 100 \\
  Number of classes & 2 & 2 & 3 & 3 & 2 & 10 \\
  \hline
  \end{tabular}
  }
  \label{tab:datasets}
\end{table}

\begin{table*}[tb]
 \caption{Average Classification Accuracy (\%) and Computation Time (sec) with Various Mean-subspace Computation Methods}
  \centering
  \resizebox{2.0\columnwidth}{!}{%
  \begin{tabular}{ccccccccccc}
  \hline
  & \multicolumn{2}{c}{Traffic} & \multicolumn{2}{c}{Car} & 
  \multicolumn{2}{c}{Wave21} & \multicolumn{2}{c}{Wave40} & 
  \multicolumn{2}{c}{Weather} \\
  \hline
  Method 
  & Accuracy(\%) & time(sec) 
  & Accuracy(\%) & time(sec)
  & Accuracy(\%) & time(sec)
  & Accuracy(\%) & time(sec)
  & Accuracy(\%) & time(sec) \\
  \hline
  EDA &  69.00 & $1.06\times10^{2}$ & 82.50 & $2.55 \times 10^3 $ & 
  72.48 & $2.23 \times 10^1 $ & 66.85 & $2.34 \times 10^1$ & 69.77 & $1.24 \times 10^2 $ \\
  \hline
  Incremental Averaging & 68.54 & $5.85 \times 10^2$ & 85.12 & $2.38 \times 10^3$ & 
  82.69 & $8.02\times 10^0$ & 80.71 & $1.40 \times 10^1$ & 69.80 & $1.99 \times 10^1$   \\
  Karcher Mean & 69.13 & $2.62 \times 10^5$ & - & - &  
  82.34 & $1.41 \times 10^4$ & 80.95 & $2.58 \times 10^4$ & 68.48 & $1.25 \times 10^5$   \\
  ICMS & 69.94 & $5.75 \times 10^1$ & 85.52 & $5.50 \times 10^3$ &  
  82.69 & $3.19 \times 10^0$ & 80.79 & $7.58 \times 10^0$ & 69.50 & $1.62 \times 10^1$   \\
  \hline
  \end{tabular}
  }
  \label{tab:means2}
\end{table*} 

\subsection{Analysis of Various Methods on Mean Subspace Computation}
One of the characteristics in our proposed OUDA framework 
is the ICMS technique
that computes the mean-target subspace incrementally.
In this subsection, we verify that the ICMS technique
is the most efficient method for
computing the mean-target subspace
as compared to the other methods on mean-subspace computation -- 
the incremental averaging and the Karcher mean. 
The incremental averaging on a subspace simply computes the average matrix 
$\mathbf{\overline{G}}_{n}$ 
of $n$ matrices $\mathbf{G}_{i}$ 
obtained from each target mini-batch as
$\mathbf{\overline{G}}_{n}=(1-\frac{1}{n})\mathbf{\overline{G}}_{n-1}+ \frac{1}{n}\mathbf{G}_{n}$, where $\mathbf{\overline{G}}_{n-1}$ is the average matrix of previous $(n-1)$ matrices. 
The Karcher mean is the computation of the mean point 
of multiple points on a subspace. 
To validate the computation efficiency of the proposed ICMS technique, 
we evaluated the computation time of our proposed OUDA framework 
using the ICMS technique and the next-target subspace prediction 
as compared to the existing EDA method~\cite{bitarafan2016incremental}.
We also measured the classification accuracy 
and the computation time of our proposed framework, 
replacing the ICMS technique with computation of the incremental averaging or the Karcher mean. 
The metric for classification accuracy~\cite{bitarafan2016incremental} 
is the average classification accuracy for a set of mini-batches $B$:
\begin{equation}
    A(B)=\frac{\sum_{\tau =1}^{|B|} a(\tau)}{|B|},
\end{equation}
where $A(B)$ is the average classification accuracy of $|B|$ mini-batches 
and $a(\tau)$ is the accuracy for the $\tau^{th}$ mini-batch.
In this subsection, 
the feedback stage was not included in our proposed framework
since we focused on comparing the methods of mean-subspace computation.
Their comparisons of average classification accuracy and computation time 
on five datasets are shown in Table~\ref{tab:means2}.


As shown in Table \ref{tab:means2}, 
the mean-subspace computation based on the ICMS technique
reached the highest average classification accuracy 
for the Traffic, the Car and the Waveform21 datasets. 
For the Waveform40 and the Weather datasets, 
the accuracy with the ICMS computation  
was less than the accuracy of other methods 
by $0.19\%$ and $0.43\%$, respectively. 
Interestingly, the mean-subspace-based methods, 
not depending on the type of mean-subspace, 
resulted in comparable or higher average classification accuracy 
compared to the EDA method. 
For the Traffic and Weather datasets, 
the mean-subspace-based methods 
showed the average classification accuracy at least $68.54\%$ and $68.48\%$, 
respectively, which were less than the accuracy of the EDA method by $0.66\%$ and $1.85\%$, respectively. 
In the case of the Car, the Waveform21 and the Waveform40 datasets, 
the mean-subspace-based methods showed the average classification accuracy 
at least $85.12\%$, $82.34\%$ and $80.71\%$, which were higher than the accuracy of EDA by 
$3.18\%$, $13.60\%$ and $20.73\%$, respectively.
These results indicated that 
averaging the subspace leverages the performance of domain adaptation. 
The incremental averaging might be 
a reasonable alternative 
when fast computation is required 
at the cost of a small dip in recognition performance.

In terms of computation time,
our proposed ICMS method was significantly faster 
than the EDA method and the Karcher-mean computation method
for most of the datasets. 
The reason for this difference on computation time was due to the fact that
the Karcher mean was computed in an iterative process.
Since the feature dimension of the Car dataset was 4096, 
which was significantly higher 
than the feature dimension of the other datasets,
the computation time of ICMS was longer than that of the incremental averaging.
Especially, the computation time of the Karcher mean computation 
for the Car dataset was longer than 4 weeks ($2.42\times10^{6}$ sec) 
and computational burden was tremendous, 
which led to tremendous computational burden.
These results indicated that our proposed ICMS-based framework is
suitable for solving the online data classification problem.

\subsection{Effect of ICMS with Additional Components}\label{subsec:fb}
In this subsection, we investigated the performance of 
the ICMS technique combined with other components in our framework 
that considers the cumulative temporal dependency 
among the arriving target data.
Coupled with the ICMS technique,
we consider the cumulative computation of transformation matrix (Cumulative), 
the next-target-subspace prediction (NextPred),
and recursive feedback (FB) to compensate
the arriving target mini-batch $\mathbf{X}_{\mathcal{T},n}$,
the target subspace $\mathbf{P}_{\mathcal{T},n+1}$,
and the transformation matrix $\mathbf{G}_{n}$ to
the pre-aligned target mini-batch $\mathbf{X}^{pre}_{\mathcal{T},n}$,
the next-target-subspace prediction $\mathbf{P'}_{\mathcal{T},n+1}$,
and the cumulative transformation matrix $\mathbf{G}_{c,n}$, respectively.
The NextPred component is suitable for noisy target domain,
whereas the Cumulative component is beneficial for gradually evolving target domain.
Hence, it is not desirable to include both the NextPred component and the Cumulative component 
(i.e. ICMS + FB + NextPred + Cumul) in our proposed framework.
We thus selected three recursive feedback variants to combine with the ICMS technique:
\begin{enumerate}
\item[(i)]
ICMS + Cumulative~~--~~
ICMS computation and cumulative computation of the transformation matrix $\mathbf{G}_{c,n}$;
\item[(ii)]
ICMS + NextPred~~--~~
ICMS computation and the next-target-subspace prediction;
\item[(iii)]
ICMS + FB+ NextPred~~--~~
ICMS computation and the next-target-subspace prediction with recursive feedback.
\end{enumerate}

\begin{table}[!b]
 \caption{Average Classification Accuracy $A(B)$ (\%) of the ICMS Method with Recursive Feedback}
  \centering
  \begin{tabular}{ccccccc}
  \hline
  Method & Traffic & Car & Wave21 & Wave40 & Weather \\
  \hline
  ICMS+Cumulative & \textbf{71.19} & 82.5 & 81.66 & 81.54 & 69.5 \\
  ICMS+FB+NextPred & 69.28 & 83.78 & 81.9 & \textbf{82.02} & 69.48 \\
  ICMS+NextPred & 68.54 & \textbf{85.93} & \textbf{83.45} & 81.43 & \textbf{70.63} \\
  \hline
  \end{tabular}
  \label{tab:variants}
\end{table}

\begin{figure}[!h]
\centering
\subfloat[\label{fig:wave40_bar} Waveform40 dataset.]{
\includegraphics[width=0.95\linewidth]{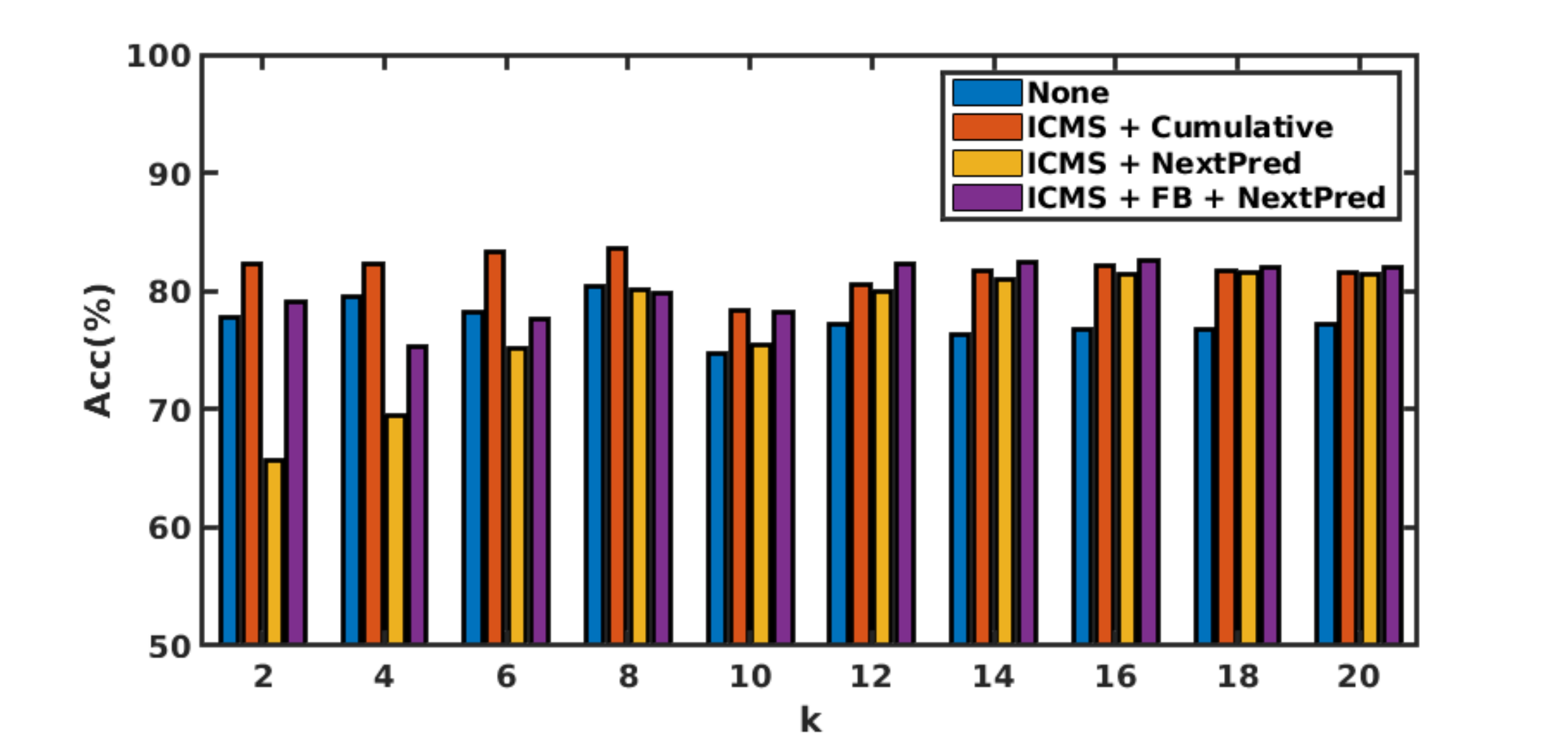}}
\\
\subfloat[\label{fig:traffic_bar} Traffic dataset.]{
\includegraphics[width=0.95\linewidth]{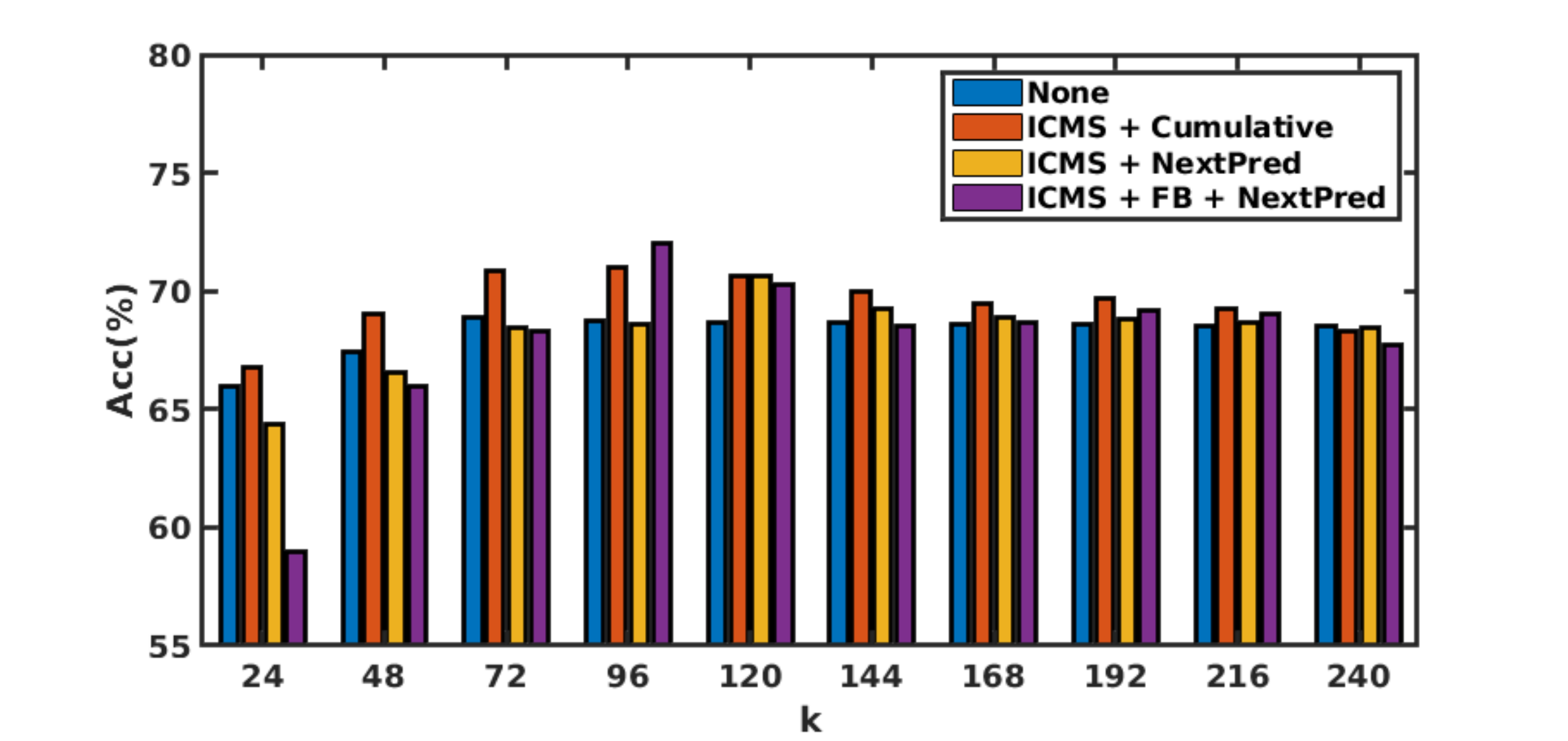}}
\caption{Effects of averaging target subspace and recursive feedback on the Waveform40 and the Traffic datasets.}
\label{fig:bars}
\end{figure}

\noindent
We then compared the average classification accuracy
on these three variants of recursive feedback 
combined with the ICMS technique.
Since the average classification accuracy depends on the feature dimension $k$,
we plotted the classification results versus the subspace dimension $k$ as shown in Fig.~\ref{fig:bars},
varying from $5\%$ to $50\%$ of the data dimension $d$. 
For the Waveform40 dataset (see Fig.~\ref{fig:bars}(a)), 
the average classification accuracy of (ICMS + FB + NextPred) outperformed 
that of other variants as the value of $k$ increased.
When the value of $k$ was small, the average classification accuracy of (ICMS + Cumulative) was the highest among other variants.
This result indicated that the effect of recursive feedback on the average classification accuracy is marginal when $k$ is small.

\begin{table*}[!ht]
 \caption{Corruption Error (\%) on Noisy CIFAR-10-C dataset}
  \centering
  \resizebox{2.0\columnwidth}{!}{%
  \begin{tabular}{cccccccccccccccccc}
  \hline
  Method & gauss & shot & impul & defoc & glass & motn & zoom &
  snow & frost & fog & brit & contr & elast & pixel & jpeg & average\\
  \hline
  ICMS~\cite{moon2020multi} & 31.73 & 27.91 & 42.29 & 19.91 & 39.09 & 21.62 & 19.20 & 23.53 & 25.52 & 
  20.60 & 12.77 & 22.12 & 31.92 & 23.54 & 30.93 & 26.18\\
  ICMS + FB & 31.94 & 28.89 & 40.95 & 18.92 & 40.39 & 21.04 & 16.56 & 22.77 & 23.54 & 19.16 & 13.11 & 20.31 & 29.53 & 23.03 & 30.47 & 25.37\\
  ICMS + NextPred & 32.42 & 28.90 & 39.41 & 17.60 & 40.62 & 21.36 & 16.78 & 21.56 & 24.08 & 19.04 & 14.42 & 19.26 & 30.00 & 23.63 & 31.44 & 25.37\\
  ICMS + FB + NextPred & 31.60 & 28.81 & 38.78 & 18.26 & 40.20 & 19.53 & 16.87 & 22.35 & 24.00 & 18.62 & 14.61 & 21.19 & 28.28 & 22.86 & 31.75 & 25.18\\
  \hline
  \end{tabular}
  }
  \label{tab:pred_effect}
\end{table*}

\begin{table*}[!ht]
 \caption{Corruption Error (\%) on CIFAR-10-C Dataset with Gradually Changing Severity Level.}
  \centering
  \resizebox{2.0\columnwidth}{!}{%
  \begin{tabular}{cccccccccccccccccccc}
  \hline
  Method & Group & Severity Level & gauss & shot & impul & defoc & glass & motn & zoom &
  snow & frost & fog & brit & contr & elast & pixel & jpeg & average\\
  \hline
  ICMS~\cite{moon2020multi} & 1 & 1,2,3,4,5 & 27.02 & 24.1 & 36.16 & 13.43 & 35.46 & 15.59 & 13.03 & 17.84 & 19.39 & 
  15.08 & 9.16 & 15.67 & 25.86 & 19.53 & 26.43 & 20.92\\
  ICMS + Cumul & 1 & 1,2,3,4,5 & 24.94 & 22.64&33.89&12.03&33.63&13.86&11.29&15.93&17.61&13.41&8.25&13.78&23.68&17.71&24.12&19.12\\
  \hline
  ICMS~\cite{moon2020multi} & 2 & 1, 3, 5 & 29.78 & 27.36 & 39.49 & 16.16 & 38.89 & 18.62 & 17.37 & 18.03 & 23.53 & 
  15.65 & 9.95 & 19.33 & 30.84 & 19.72 & 27.53 & 23.48\\
  ICMS + Cumul & 2 & 1, 3, 5 & 29.05 & 24.15 & 38.44 & 15.61 & 35.69 & 18.79 & 13.23 & 21.62 & 20.71 & 18.03 & 10.59 & 14.50 & 25.45 & 20.91 & 27.28 & 22.27\\
  \hline
  \end{tabular}
  }
  \label{tab:cumul_effect}
\end{table*}

\begin{table*}[!ht]
 \caption{Corruption Error (\%) on CIFAR-10-C dataset for non-NN-based methods (severity level 5).}
  \centering
  \resizebox{2.0\columnwidth}{!}{%
  \begin{tabular}{ccccccccccccccccccc}
  \hline
  Method & Classifier & gauss & shot & impul & defoc & glass & motn & zoom &
  snow & frost & fog & brit & contr & elast & pixel & jpeg & average\\
  \hline
  Source (No Adaptation)& NN & 72.33 & 65.7 & 72.9 & 46.9 & 54.3 & 34.8 & 42.0 & 25.1 & 41.3 & 
  26.0 & 9.3 & 46.7 & 26.6 & 58.5 & 30.3 & 43.51\\
  \hline
  First Target & SVM & 29.50 & 28.71&39.00&18.49&39.63&18.42&16.71&21.42&22.00&19.39&12.37&17.69&29.27&21.62&29.94&24.28\\
    ICMS-SVM & SVM & 26.95 &24.04&36.07&13.46&35.47&15.40&12.89&17.78&19.21&14.98&9.18&15.56&25.67&19.46&26.38&20.83\\
  \hline
  \end{tabular}
  }
  \label{tab:non-nn}
\end{table*}

For the Traffic dataset (see Fig.~\ref{fig:bars}(b)), 
the average classification accuracy of (ICMS + Cumulative) 
was higher than that of other variants for most $k$ values.
However, the average classification accuracy of (ICMS + FB + NextPred) 
was the highest for $k=96$, 
and this accuracy was the highest among the accuracies of all the variants.
This result showed that recursive feedback significantly 
leverages the classification performance 
with proper subspace dimension $k$.

Table~\ref{tab:variants} showed the average classification accuracy 
for the ICMS method coupled with 3 recursive feedback variants of 
our proposed framework when the values of subspace dimension $k$ 
were set to the default values in Table~\ref{tab:datasets}.
For the Traffic dataset, 
the average classification accuracy of (ICMS + Cumulative) was the highest among the 3 variants of our proposed framework.
This result was shown in Fig.~\ref{fig:bars}(b), 
where the average classification accuracy of (ICMS + FB + NextPred) 
decreased after the value of $k=96$.
The average classification accuracy of (ICMS + NextPred) 
was the highest ($71.19\%$) for the Waveform40 dataset.
This result was shown in Fig.~\ref{fig:bars}(a) for $k=20$.

For the Car, Wave21, and Weather datasets, 
the average classification accuracies of (ICMS + FB + NextPred) 
were lower than that of (ICMS + NextPred).
The reason for these results is that adjacent target mini-batches 
lacked strong temporal dependency (the Car dataset) 
and the subspace dimension $k$ was too small 
(the Wave21 and Weather datasets).
Hence, we concluded that the recursive feedback stage 
improves the classification accuracy 
when the subspace dimension is large enough 
and the target mini-batches have strong temporal dependency.

\subsection{Effect of Next-Target-Subspace Prediction and Recursive Feedback}
In this subsection, we demonstrated 
the effect of next-target-subspace prediction and recursive feedback 
in our proposed framework.
To verify the robustness of subspace prediction and recursive feedback,
we measured the corruption errors on four variants for comparison -- 
ICMS, ICMS + FB, ICMS + NextPred, ICMS + FB + NextPred.
To conduct experiments on a large-scale dataset with noisy domain, 
we blended several corruption levels of images in CIFAR-10-C dataset.
Specifically, we replaced 20\% of the target images of corruption level 5 with corruption level 1.
As shown in Table~\ref{tab:pred_effect},
the errors of (ICMS + FB) and (ICMS + NextPred) 
showed lower corruption errors than ICMS~\cite{moon2020multi},
and the error of (ICMS + FB + NextPred) was the lowest among all the variants.
From these results, we concluded that
subspace prediction and recursive feedback stages 
increase the robustness of our OUDA framework 
when the target subspaces are noisy.

\subsection{Effect of Cumulative Computation of Transformation Matrix}
Transformation matrix computed by our proposed framework 
aligns target data closer to the source domain.
Especially, the transformation matrix obtained by cumulative computation
showed significant improvement for gradually evolving target domains.
To validate the effectiveness of cumulative computation,
we measured the classification error of images in CIFAR-10-C dataset,
changing the corruption level from 1 to 5.
To compare the results on different evolving velocity of the target domain,
we set the increment of corruption level as 1 for one group and 2 for another.
As shown in Table~\ref{tab:cumul_effect},
the cumulative computation of transformation matrix component lowered 
the corruption error from 20.92\% to 19.12\% and 23.48\% to 22.27\% in group 1 and group 2, respectively.
From these results, we concluded that
cumulative computation of transformation matrix is effective 
when the source and the first target is similar 
and the target domain is gradually changing.

\subsection{Classifier Update for Non-Neural-Network-Based Methods}
In this subsection, we verified the effect of non-NN-based adaptive classifier on our proposed framework. 
For fair comparisons, we measured the average classification accuracy 
of variants of our framework (described in Section~\ref{subsec:fb})
with and without classifier adaptation.
The number of source data was 10\% of the number of entire data sample and the batch-size of the target data was $N_{\mathcal{T}}=2$.
The classifier was updated by each target mini-batch with predicted pseudo-labels.
Tables~\ref{tab:adapt-traffic} and \ref{tab:adapt-weather} showed the average classification accuracy of our framework for the Traffic and the Weather datasets, respectively.
For all the variants of our proposed ICMS method, 
the classification accuracy of adaptive classifier was higher, 
from 0.21\% to 3.88\%, than that of the source classifier.
These results showed that the adaptive classifier leveraged the performance of our proposed framework.

Table~\ref{tab:non-nn} showed the effect of 
adaptive SVM classifier for the CIFAR-10-C dataset. 
Our method (ICMS-SVM) improved the test-time adaptation 
compared to the classifier trained on the first target mini-batch. 
For all the corruption types, errors of the ICMS-SVM method were lower than that of the first target, 
from 7.51\% to 27.20\%. 
These results showed that the ICMS technique improves the 
online adaptation for large-scale datasets.

\begin{table}[!ht]
 \caption{Average Classification Accuracy $A(B)$ (\%) of the ICMS Method with Adaptive Classifier on Traffic Dataset}
  \centering
  \begin{tabular}{ccc}
  \hline
  Method & Fixed Classifier & Adaptive Classifier \\
  \hline
  ICMS + NextPred & 67.89 & 70.63  \\
  ICMS + Cumulative & 69.77 & 71.63 \\
  ICMS + FB + NextPred & 70.63 & 71.44 \\
  \hline
  \end{tabular}
  \label{tab:adapt-traffic}
\end{table}

\begin{table}[!ht]
 \caption{Average Classification Accuracy $A(B)$ (\%) of the ICMS Method with Adaptive Classifier on Weather Dataset}
  \centering
  \begin{tabular}{ccc}
  \hline
  Method & Fixed Classifier & Adaptive Classifier \\
  \hline
  ICMS + NextPred & 71.68 & 72.09  \\
  ICMS + Cumulative & 72.07 & 72.24 \\
  ICMS + FB + NextPred & 71.89 & 72.04 \\
  \hline
  \end{tabular}
  \label{tab:adapt-weather}
\end{table}

\subsection{Online Classification Accuracy of Proposed Framework}\label{sub:result_online}
\begin{figure}[!ht] 
    \centering
  \subfloat[\label{fig:traffic} Traffic dataset]{%
      \includegraphics[width=0.5\linewidth]{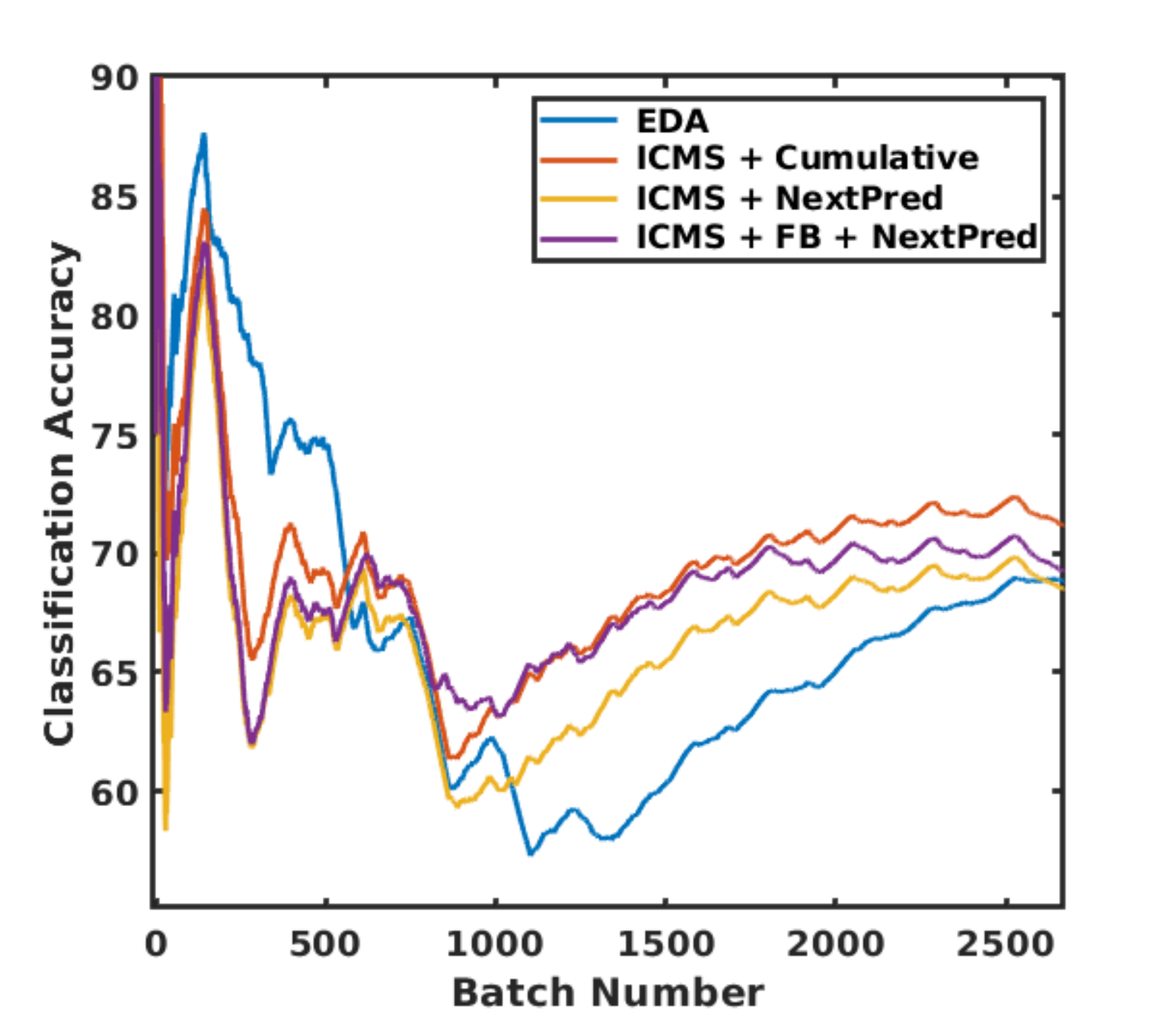}}
    \hfill
  \subfloat[\label{fig:car} Car dataset]{%
        \includegraphics[width=0.5\linewidth]{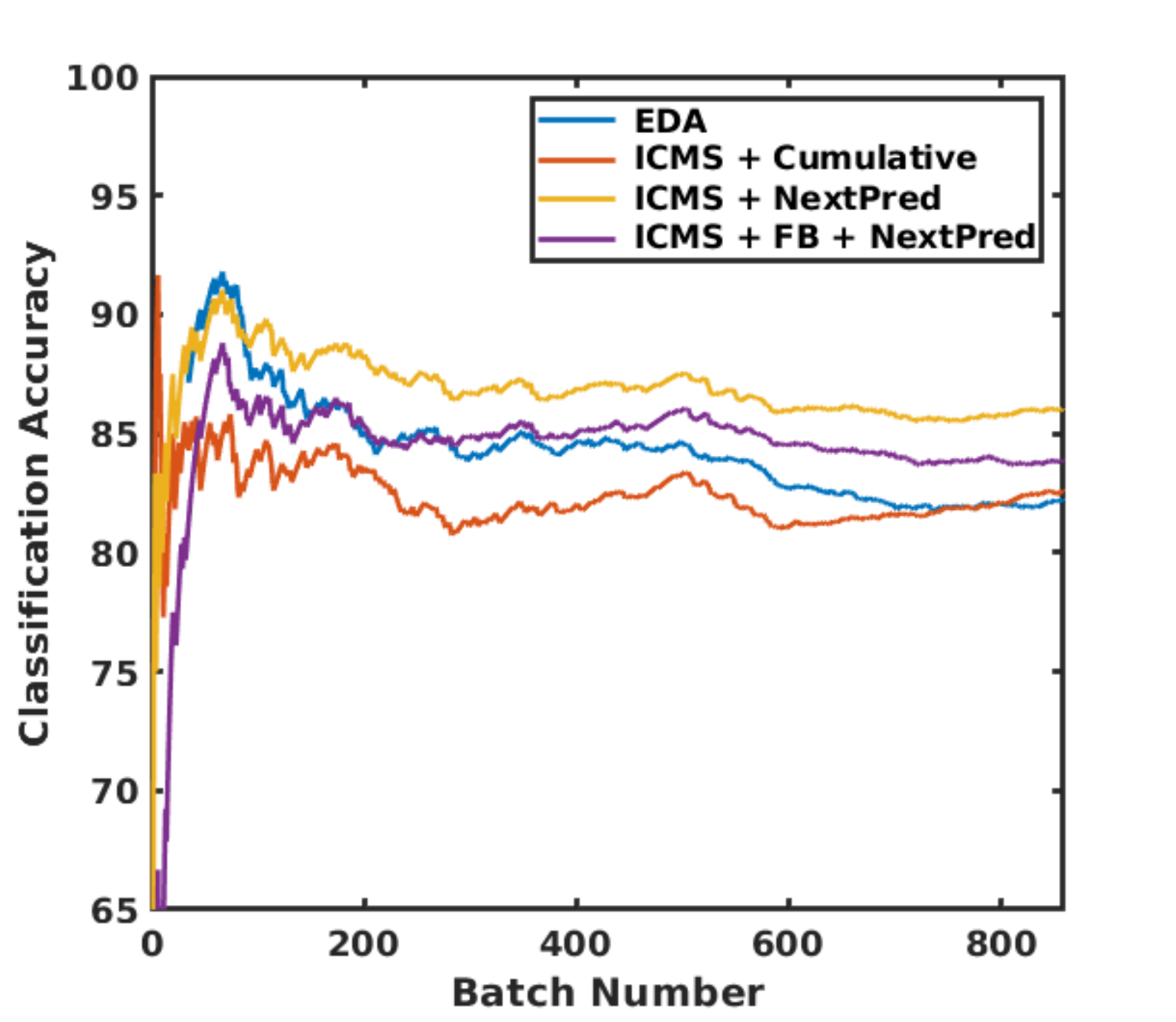}}
    \hfill
    \\
    \centering
  \subfloat[\label{fig:wave21} Waveform21 dataset]{%
        \includegraphics[width=0.5\linewidth]{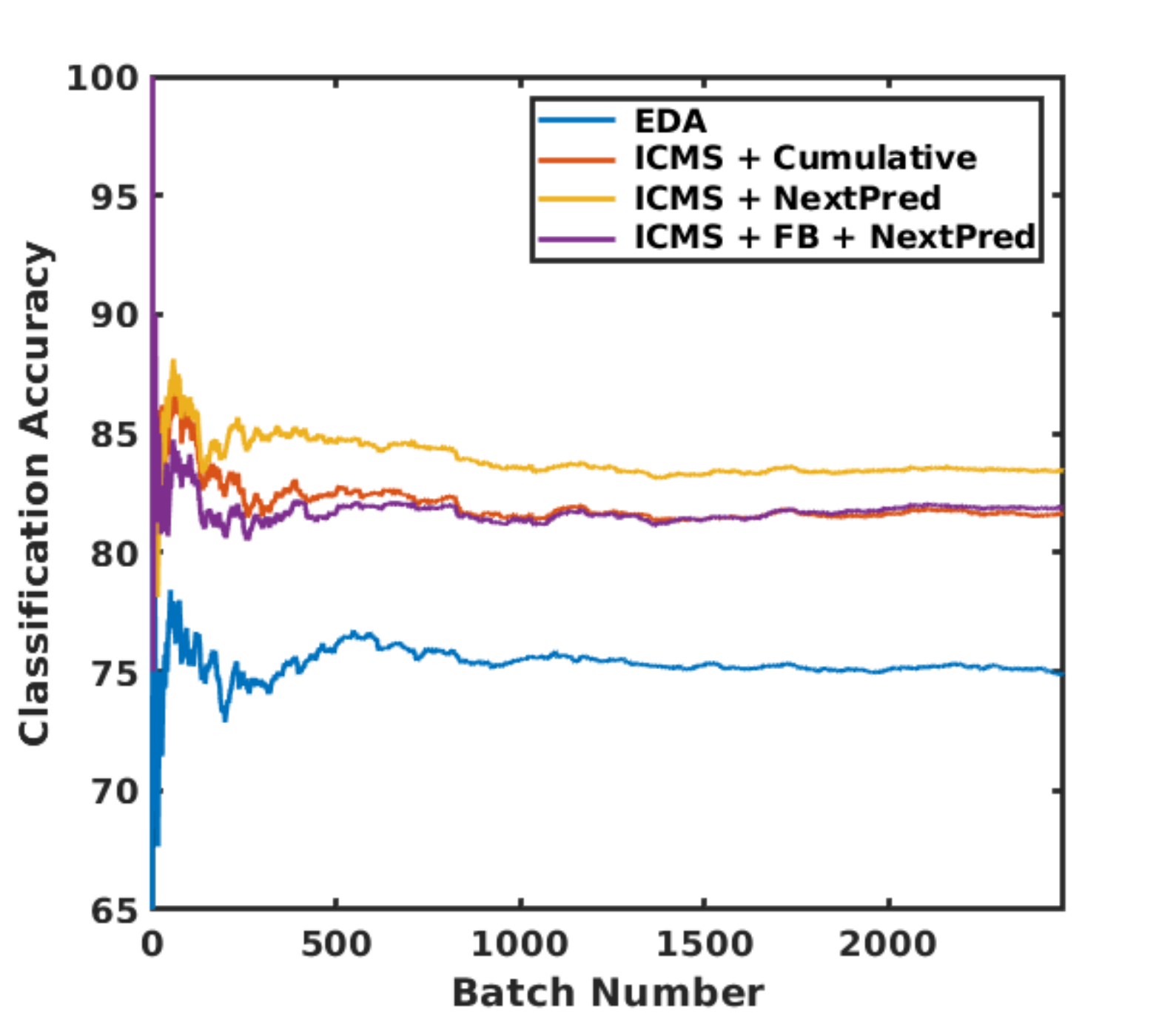}}
    \hfill
  \subfloat[\label{fig:wave40} Waveform40 dataset]{%
        \includegraphics[width=0.5\linewidth]{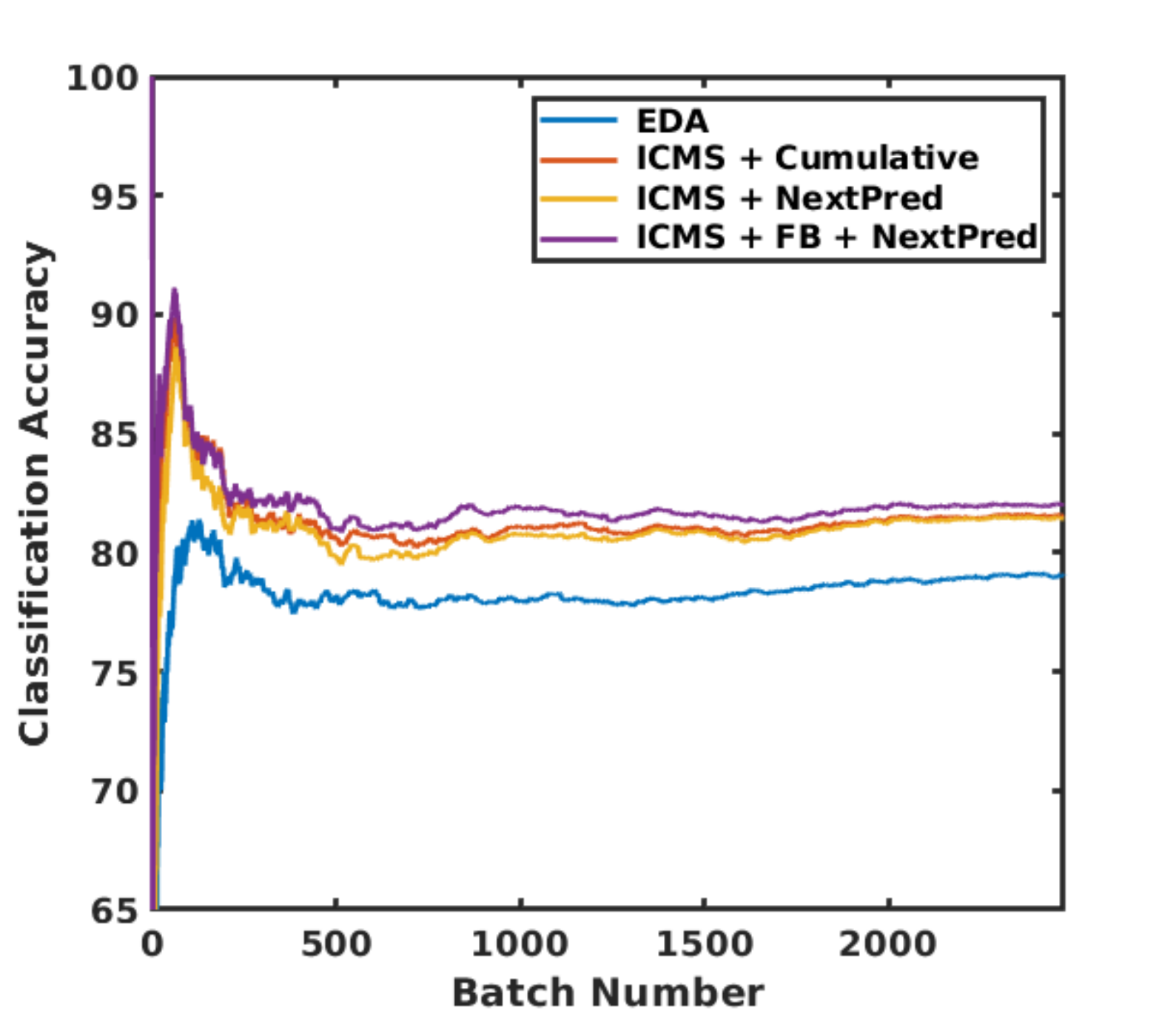}}
    \hfill
  \caption{Classification accuracy of each mini-batch $a(\tau)$ (\%) computed by the baseline and variants of the proposed framework.}
  \label{fig:plot} 
\end{figure}

To further understand the characteristics of our proposed OUDA framework for online classification performance,
we investigated the effect of each stage toward the performance 
of proposed framework.
For consistency, we selected the same ICMS method combined with 
3 recursive feedback variants as in the previous subsection.
In addition to the average accuracy $A(B)$ metric, 
we also compared the classification accuracy
as each target mini-batch arrived 
by plotting $a(\tau)$ versus the number of mini-batches $|B|$ 
(see Fig.~\ref{fig:plot}).
Figure~\ref{fig:plot} indicated that 
our proposed framework and its variants 
outperformed the EDA method for most of the datasets. 
In Fig.~\ref{fig:plot}(a), a sudden drift occurred from 
the $100^{th}$ mini-batch to the $900^{th}$ mini-batch, 
which resulted in an abrupt decrease of the accuracy.
After the $900^{th}$ mini-batch, 
the accuracy was recovered as the number of arriving mini-batch increased.
For the Car dataset, the average accuracy was slightly decreased 
since the target data were evolving in a long time-horizon 
(i.e., from 1950 to 1999), 
which resulted in more discrepancies 
between the source and the target domains.
In terms of the gap between the highest and the lowest accuracies,
the Traffic and Waveform40 datasets showed more gap (15\% to 25\%) 
compared to other datasets (5\% to 10\%).
We concluded that the computation of transformation matrix (Cumulative) 
and recursive feedback (FB) showed more significant effect 
for dynamic datasets.

\subsection{Convergence of Computed Mean-target Subspace}
\begin{figure}[!ht]
\centering
\subfloat[$d(\mathbf{P}_{\mathcal{S}}, \mathbf{\overline{P}}_{\mathcal{T},n})$\label{fig:wave21_PsPtmean}]{
\includegraphics[width=0.45\linewidth]{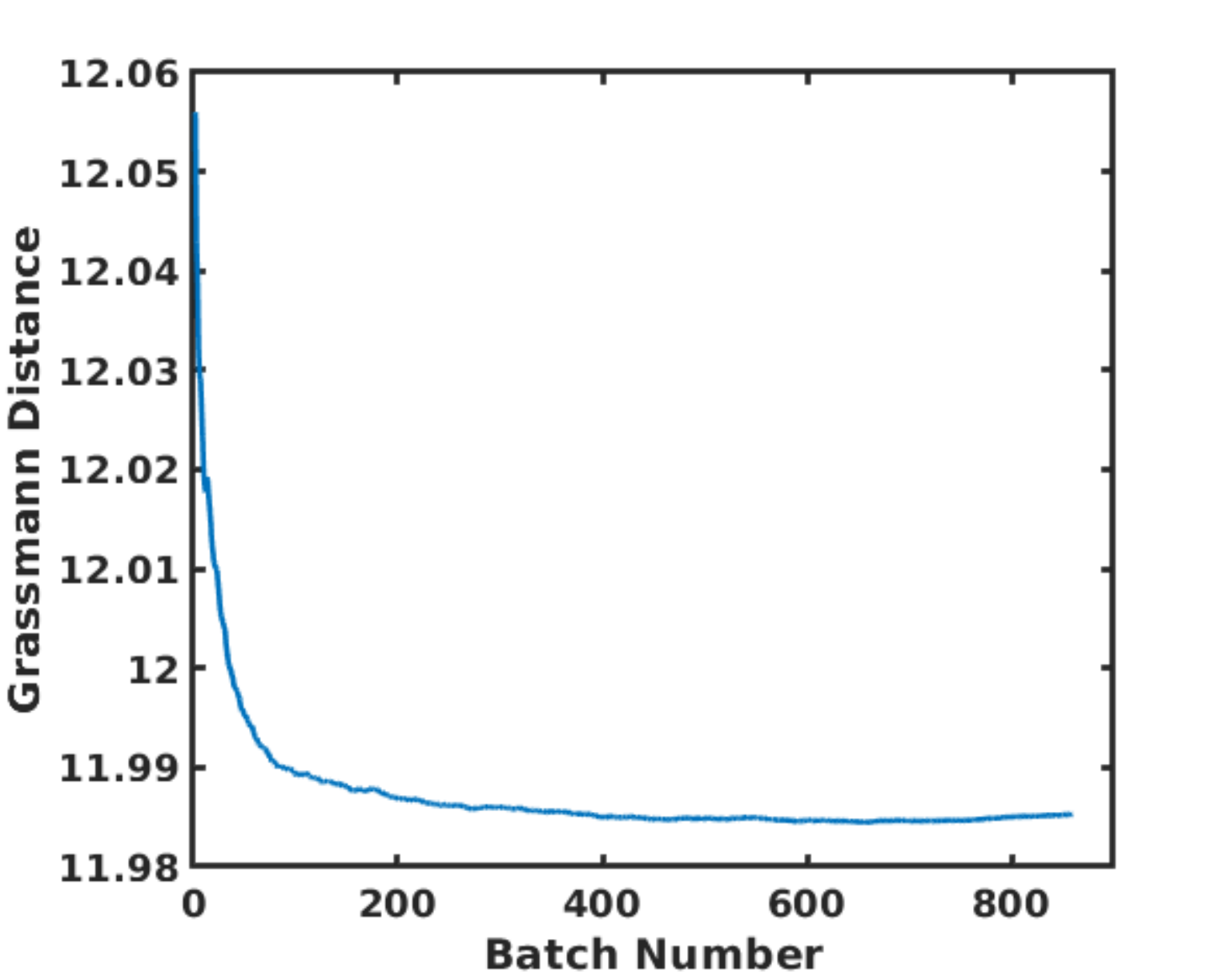}}
\subfloat[$d(\mathbf{\overline{P}}_{\mathcal{T},n-1}, \mathbf{\overline{P}}_{\mathcal{T},n})$\label{fig:wave21_PtprevPt}]{
\includegraphics[width=0.45\linewidth]{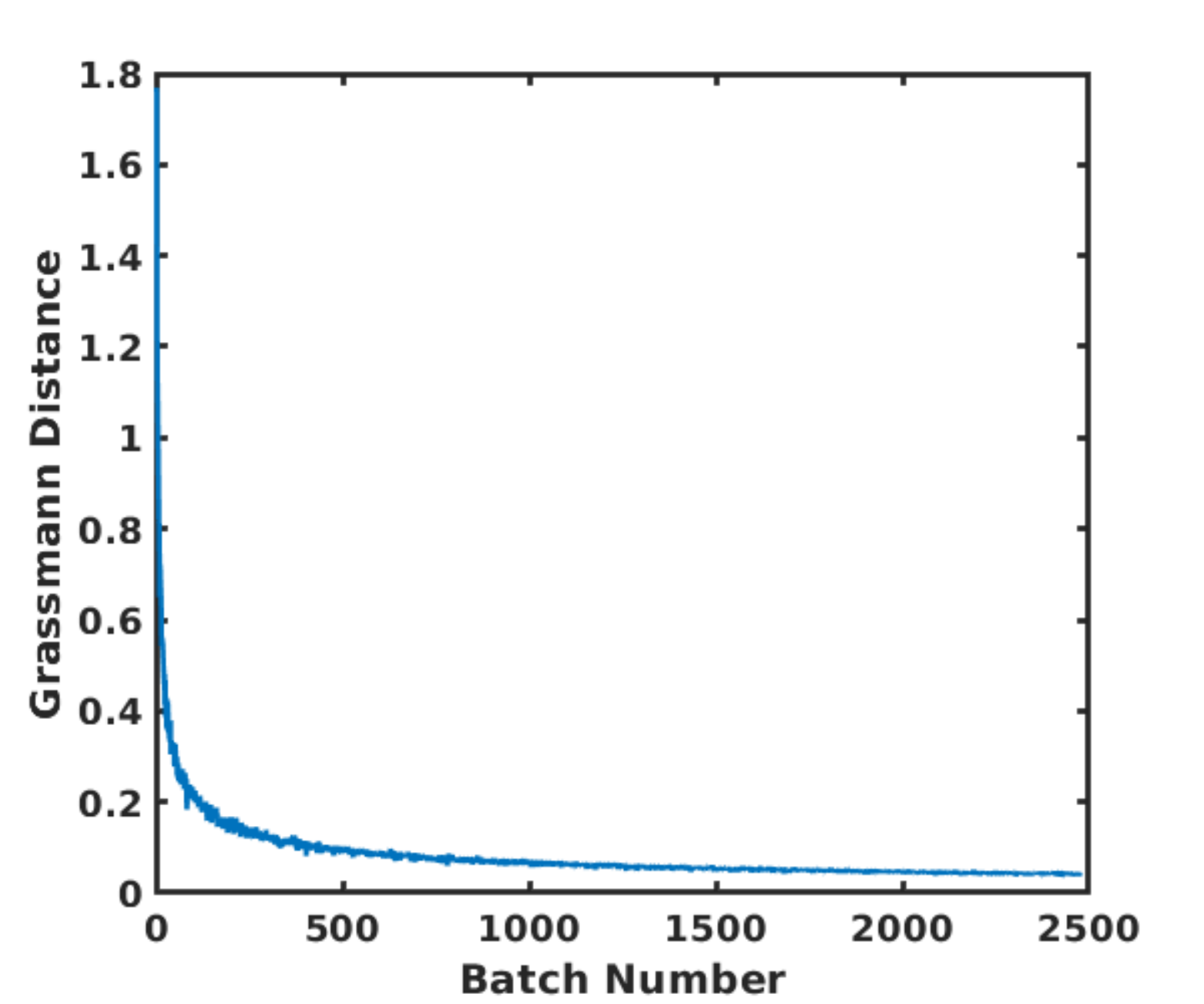}}\\
\subfloat[$d(\mathbf{P}_{\mathcal{S}}, \mathbf{\overline{P}}_{\mathcal{T},n})$\label{fig:car_PsPtmean}]{
\includegraphics[width=0.45\linewidth]{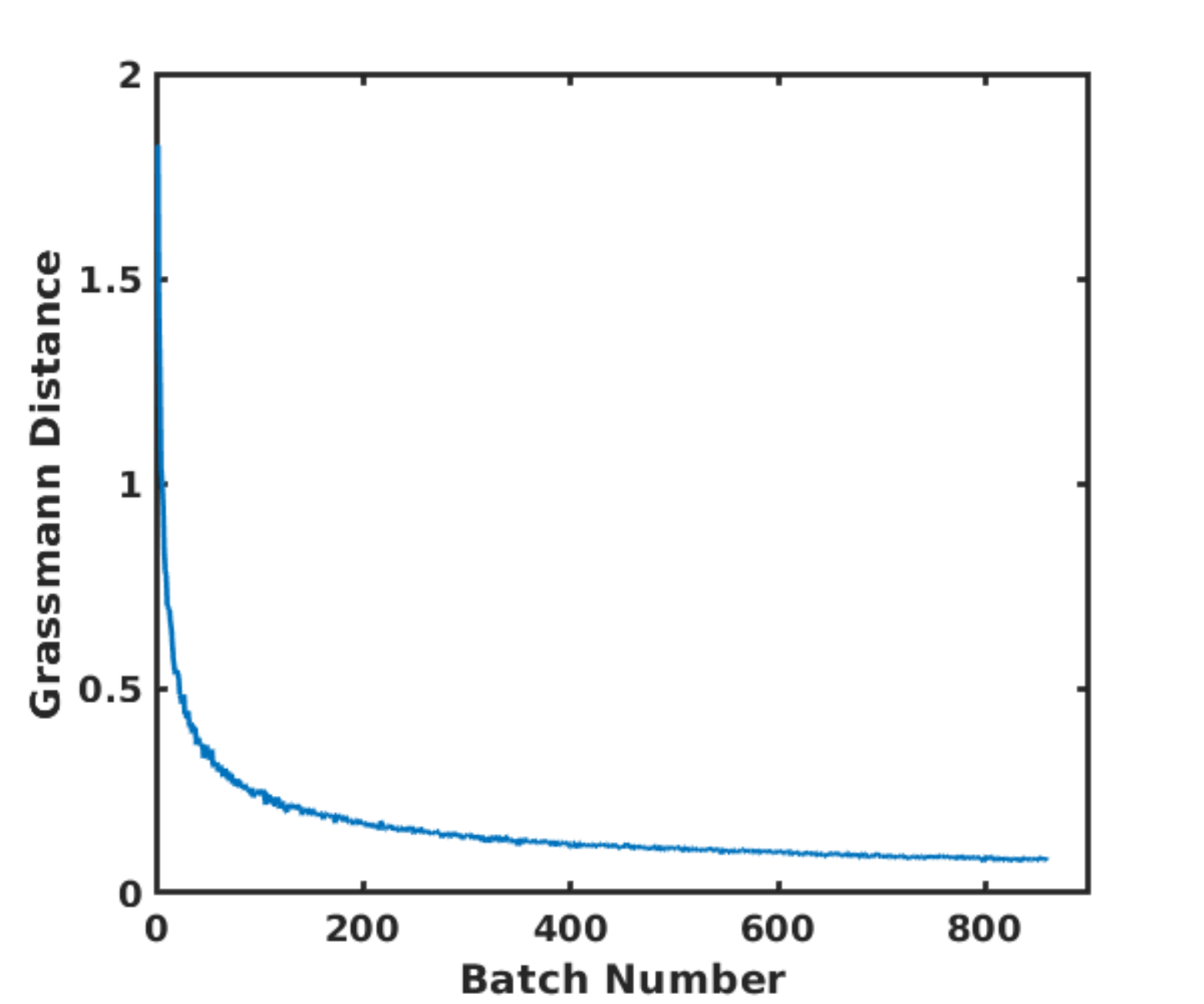}}
\subfloat[$d(\mathbf{\overline{P}}_{\mathcal{T},n-1}, \mathbf{\overline{P}}_{\mathcal{T},n})$\label{fig:car_PtprevPt}]{
\includegraphics[width=0.45\linewidth]{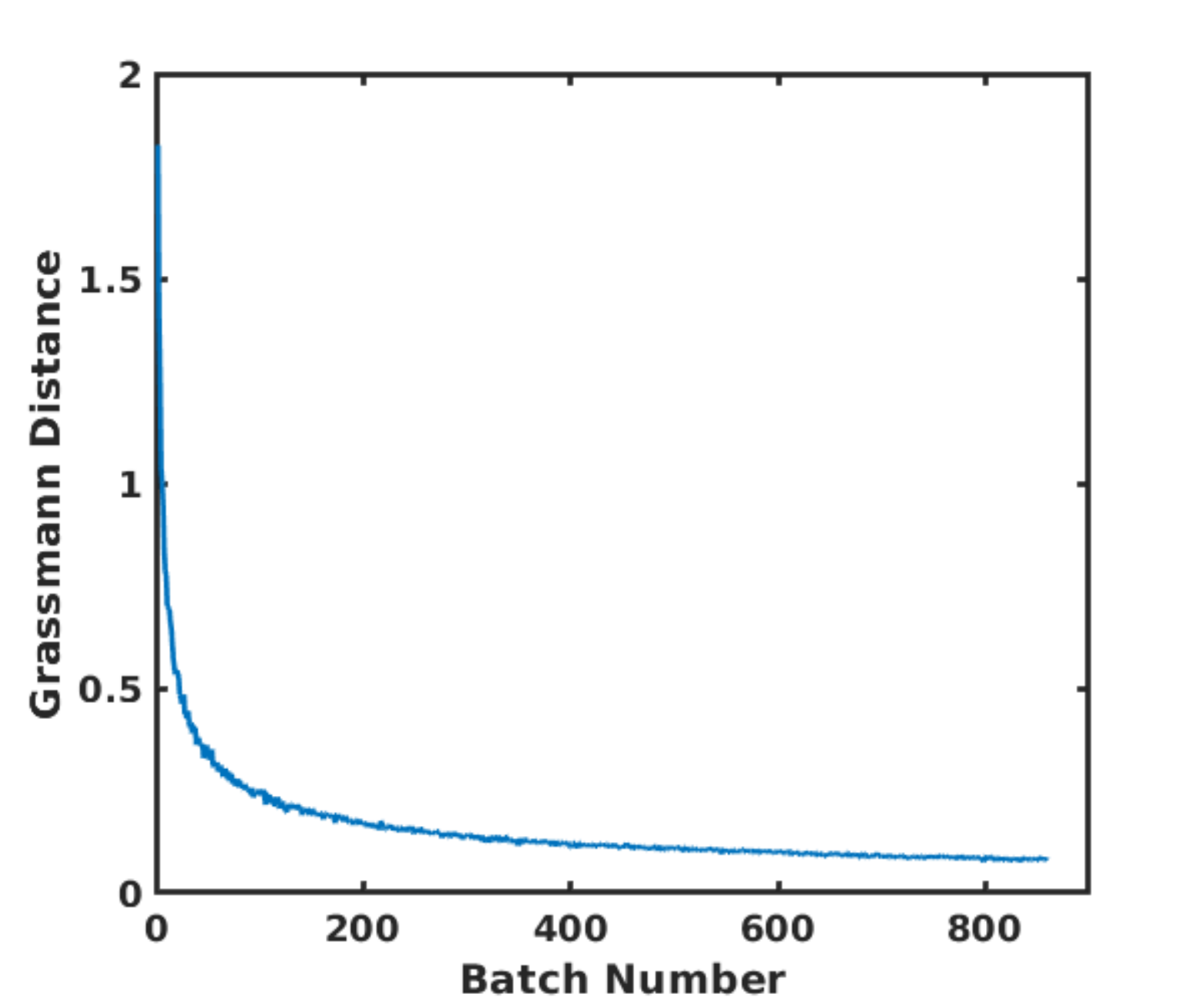}}
\caption{Geodesic distances between two subspaces. Figures (a) and (b) used the Waveform21 dataset. Figures (c) and (d) used the Car dataset.}
\label{fig:convergence}%
\end{figure}

In this subsection, we empirically demonstrated the convergence of 
mean-target subspace computed by the ICMS technique.
To prove that the mean-target subspace converges to a fixed subspace,
we computed and plotted the geodesic distance 
between the source subspace $\mathbf{P}_{\mathcal{S}}$ 
and the mean-target subspace $\mathbf{\overline{P}}_{\mathcal{T},n}$ 
(i.e., $d(\mathbf{P}_{\mathcal{S}}, \mathbf{\overline{P}}_{\mathcal{T},n})$).
Furthermore, we observed the stability of ICMS technique 
by showing that the geodesic distance~\cite{hamm2008grassmann} between
two consecutive mean-target subspaces $\mathbf{\overline{P}}_{\mathcal{T},n-1}$ and $\mathbf{\overline{P}}_{\mathcal{T},n}$ (i.e., $d(\mathbf{\overline{P}}_{\mathcal{T},n-1}, \mathbf{\overline{P}}_{\mathcal{T},n})$) converged to zero
as the number of arriving target mini-batches increased.
Figures~\ref{fig:convergence}(a) and~\ref{fig:convergence}(c) showed the geodesic distances 
between the source subspace $\mathbf{P}_{\mathcal{S}}$ 
and the mean-target subspace $\mathbf{\overline{P}}_{\mathcal{T},n}$ 
for the Waveform21 and the Car datasets, respectively.
Their geodesic distances converged to a constant.
Similarly, Figures~\ref{fig:convergence}(b) 
and \ref{fig:convergence}(d) showed the geodesic distances  
between the two consecutive mean-target subspaces 
$\mathbf{\overline{P}}_{\mathcal{T},n-1}$ 
and $\mathbf{\overline{P}}_{\mathcal{T},n}$ 
for the Waveform21 and the Car datasets, respectively.
These geodesic distances converged to zero.
Thus, the convergence of the above geodesic distances 
validated the proposed ICMS technique  for computing
the mean-target subspace.
\vspace{-0.1in}

\subsection{Parameter Sensitivity}
\begin{figure}[ht]
\centering
\subfloat[\label{fig:traffic_3d}]{
\includegraphics[width=0.5\linewidth]{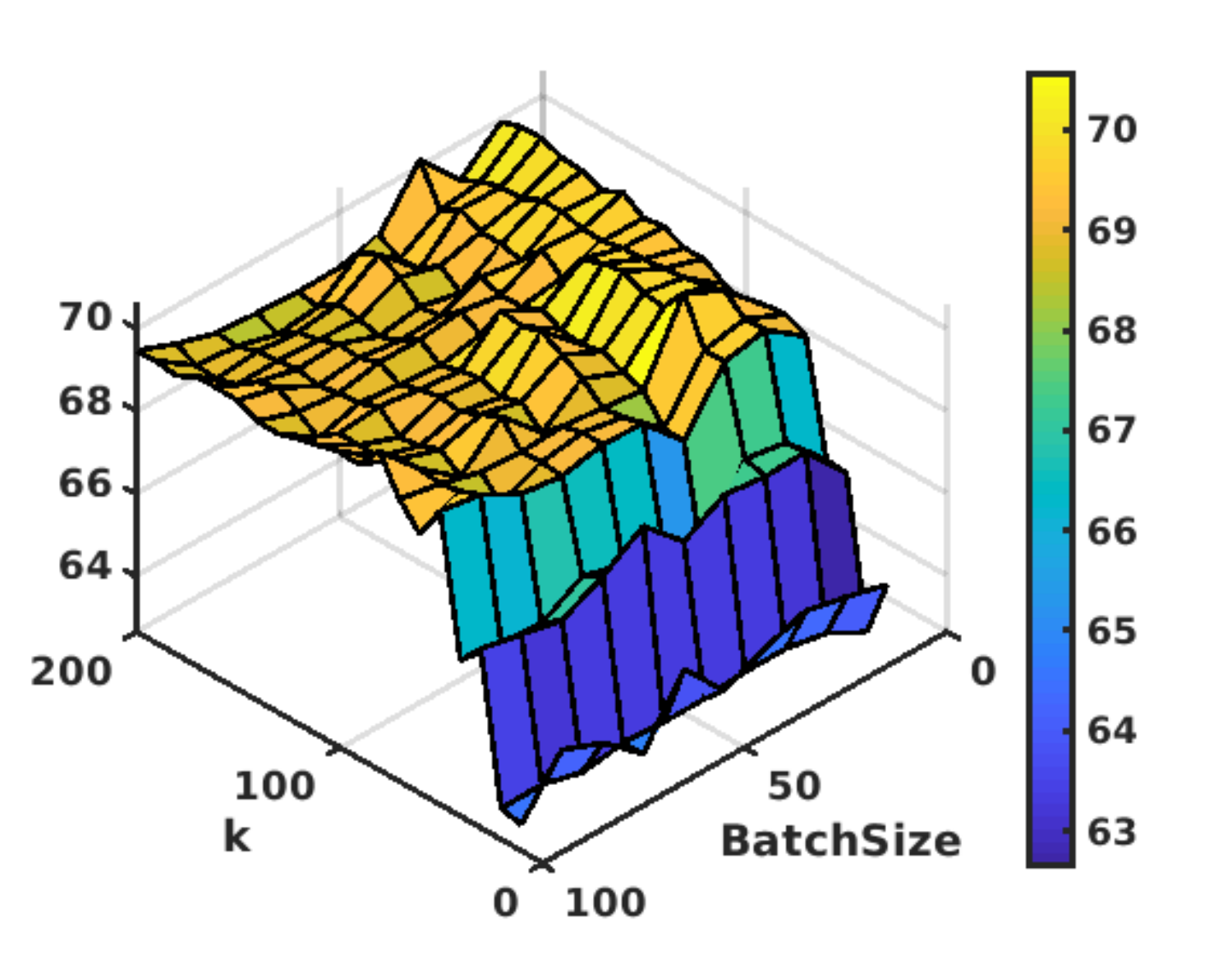}}
\subfloat[\label{fig:car_3d}]{
\includegraphics[width=0.5\linewidth]{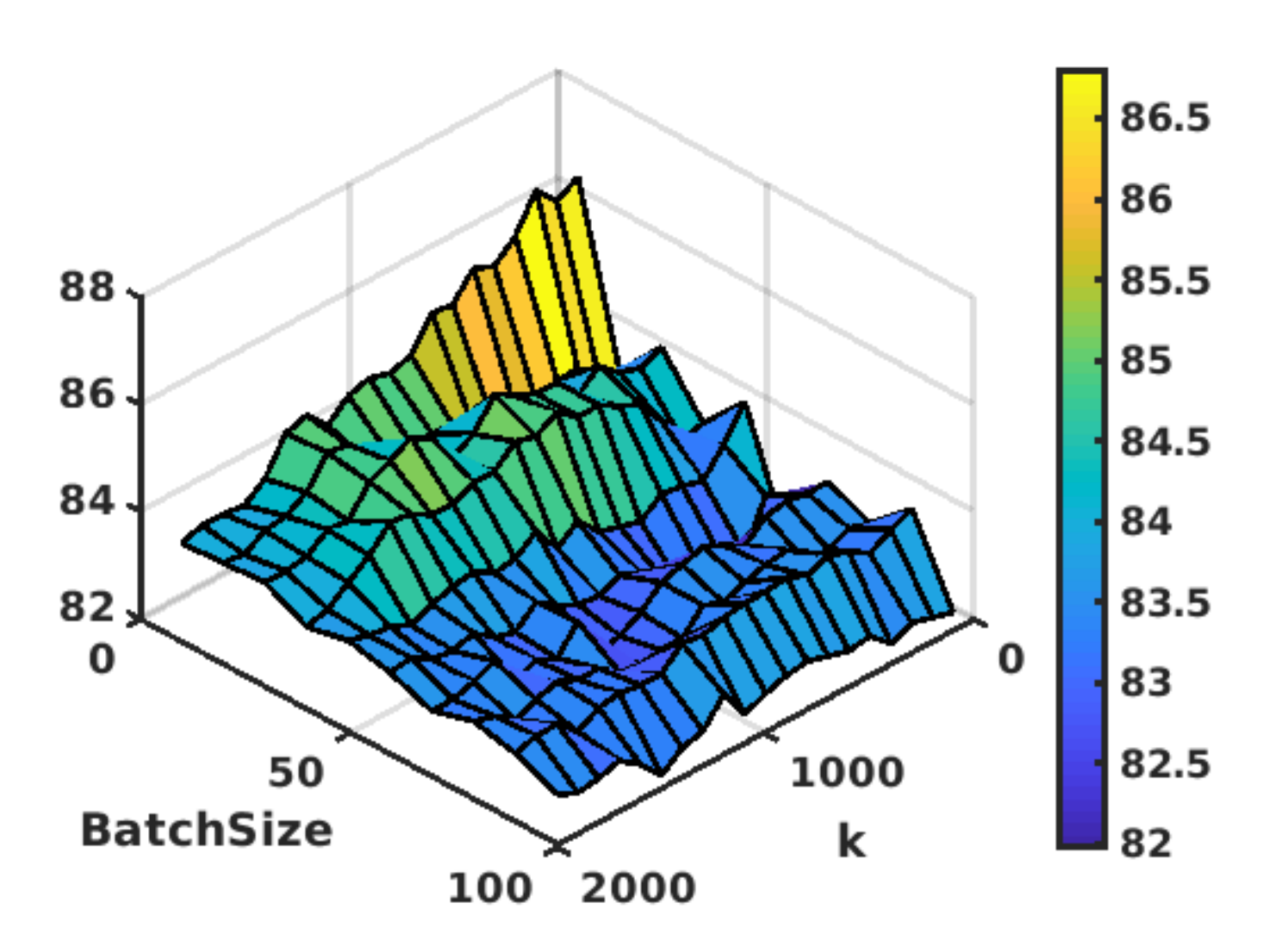}}
\caption{Sensitivity analysis on $k$ and batchsize on (a) Traffic (b) Car dataset.}%
\label{fig:3ds}
\end{figure}

We also visualized the effect of both parameters $k$ and $N_{\mathcal{T}}$
by measuring the accuracy corresponding to their various values. 
Figures~\ref{fig:3ds}(a) 
showed that the classification accuracy 
significantly depended on the value of $k$ 
while it remained relatively stable with the values of $N_{\mathcal{T}}$. 
From Fig.~\ref{fig:3ds}(b), the classification accuracy was the highest 
when $k$ and $N_{\mathcal{T}}$ were small, 
but the difference between the highest and the lowest accuracies was $4.5\%$, 
which was relatively negligible compared 
to Fig.~\ref{fig:3ds}(a). 
\vspace{-0.1in}

\subsection{Comparison with Existing Manifold-based Traditional Methods}
We compared the {\em average} classification accuracy 
of the proposed OUDA framework with two existing manifold-based
traditional methods~--~Evolving Domain Adaptation (EDA)~\cite{bitarafan2016incremental} 
and Continuous Manifold Alignment (CMA)~\cite{hoffman2014continuous}.
The CMA method has two variations depending on the domain adaptation techniques~--~
GFK~\cite{gong2012geodesic} and SA~\cite{fernando2013unsupervised}.
Among the variants of our proposed OUDA framework,
we selected the (ICMS + FB + NextPred) variant for comparison 
with EDA and CMA methods. 
This (ICMS + FB + NextPred) variant uniquely shows major contributions 
of  the ICMS technique combined with recursive feedback in our proposed framework.
In the comparison, 
we assigned the parameter values of 
the (ICMS + FB + NextPred) variant the same as 
those of EDA and CMA methods. 
The batch-size of the arriving target data was $N_{\mathcal{T}}=2$. 
Except for the EDA method that adopted 
the Incremental Semi-Supervised Learning (ISSL) technique 
for classifying the unlabeled target data, 
all other approaches adopted the basic 
Support-Vector-Machine~\cite{suykens1999least} classifiers
for target-label prediction. 

The results of comparison on the five datasets 
are shown in Table~\ref{tab:shallow}.
In Table~\ref{tab:shallow}, 
our proposed OUDA framework  
obtained a higher average classification accuracy than traditional methods 
(i.e., CMA+GFK, CMA+SA, and EDA methods) for all the five datasets. 
For these five datasets, the highest average classification accuracy 
among the traditional methods were 
$69.00\%$, $82.73\%$, $74.65\%$, $79.66\%$, $64.79\%$.
The average classification accuracy of 
our proposed OUDA framework for the respective five datasets were 
$69.28\%$, $83.78\%$, $81.9\%$, $82.02\%$, and $69.48\%$. 
This result demonstrated that our proposed OUDA framework  
outperformed the other traditional methods 
on the data classification problem for all the datasets.

\begin{table}[!htb]
 \caption{Overall Accuracy $A(B)$ (\%) of the Traditional Methods and the Proposed OUDA Framework}
  \centering
  \begin{tabular}{ccccccc}
  \hline
  Method & Traffic & Car & Wave21 & Wave40 & Weather \\
  \hline
  CMA+SA & 41.33 & 56.45 & 33.84 & 33.05 & 31.40\\
   CMA+GFK \cite{bitarafan2016incremental, hoffman2014continuous} & 68.87 & 82.73 & 69.15 & 68.77 & 63.81 \\
  EDA \cite{bitarafan2016incremental} & 69.00 & 82.59 & 74.65 & 79.66 & 64.79 \\
  \hline
  Proposed Framework & \textbf{69.28} & \textbf{83.78} & \textbf{81.9} & \textbf{82.02} & \textbf{69.48} \\
  \hline
  \end{tabular}
  \label{tab:shallow}
\end{table}

\subsection{Comparison with Neural-Network-based Methods}
We compared the corruption error (\%) with several NN-based-online-adaptation 
methods -- classifier with no adaptation with target data (Source),
Test Entropy Minimization (Tent)~\cite{DBLP:conf/iclr/WangSLOD21},
and Test-Time Normalization (BN)~\cite{schneider2020improving}.
To utilize our framework on test-time adaptation tasks,
we first extracted the features before the classifier module of NN. 
Our proposed framework then transforms these features
and input the transformed features through the classifier.
Since the source data are not accessible 
in test-time adaptation tasks, 
our framework aligns the arriving target data 
to the initial target domain instead of the source domain.
For the experiments on corruption, 
we used Wide Residual Network with 28 layers (WRN-28-10)~\cite{BMVC2016_87} 
and Residual Network with 26 layers (ResNet26)~\cite{he2016deep}  
on the CIFAR-10-C dataset.
Dimensions of the features extracted from WRN-28-10 and ResNet26 were 640 and 256, respectively.
These extracted features were transformed by our proposed framework.
We optimized the parameters of WRN-28-10 with Adam optimizer~\cite{DBLP:journals/corr/KingmaB14}, 
where the batchsize and learning rate were 200 and 0.001, respectively.
For ResNet26, the batchsize and learning rate were 128 and 0.001, respectively.
The classification errors for various types of corruption are shown in Table~\ref{tab:nn_avg}. 

For both WRN-28-10 and ResNet26, our ICMS-NN showed the lowest error among the NN-based methods.
For WRN-28-10 and ResNet26, the corruption errors of ICMS-NN were 18.24\% and 14.1\%, respectively.
In terms of the difference of errors, our ICMS was more effective for non-NN models than NN models.
This is due to the characteristic of ICMS, 
which computes the transformation matrix based on the GFK technique.
However, the errors of NN-based methods (BN, Tent, and ICMS-NN) 
were lower than the non-NN-based methods (First Target, ICMS-SVM) for most of the corruption types, 
which indicate that the NN-based models adapted with target data better than the non-NN-based models.
These results showed that our proposed framework leverages the test-time-adaptation task for
both NN-based and non-NN-based models.

\begin{table}[!htb]
 \caption{Average Corruption Error (\%) on CIRAR-10-C Dataset for NN-based Methods (severity level 5)}
  \centering
  \begin{tabular}{ccc}
  \hline
  Method & WRN-28-10 & ResNet26 \\
  \hline
  Source & 43.51 & 40.8 \\
   BN \cite{schneider2020improving} & 20.4 & 17.3  \\
  Tent \cite{DBLP:conf/iclr/WangSLOD21} & 18.59 & 14.3 \\
  \hline
  ICMS-NN & \textbf{18.24} & \textbf{14.1} \\
  \hline
  \end{tabular}
  \label{tab:nn_avg}
\end{table}

\section{Conclusion and Future Work}
We have proposed a multi-stage framework 
for tackling the OUDA problem,
which includes a novel technique of
incrementally computing the mean-target subspace
on a Grassmann manifold.
We have proved that
the mean-target subspace computed by the ICMS method 
is a valid close approximation to the Karcher mean
with efficient computation time.
To achieve more robust online domain adaptation,
we proposed to utilize subspace prediction and 
cumulative computation of transformation matrix
by considering the flow of target subspaces 
on the Grassmann manifold.
We also verified that the adaptive classifier 
improves the performance of online adaptation.
Extensive experiments on various datasets demonstrated that 
our proposed OUDA framework outperforms 
existing traditional manifold-based methods 
and NN-based learning methods
in terms of classification performance and computation time.
Moreover, contribution of each stage 
in our proposed framework has been analyzed
by comparing the variants of our proposed method
with or without each stage.
Future work includes extension of the proposed ICMS technique 
and the OUDA framework to
more challenging domain adaptation applications.
\vspace{-0.05in}

\ifCLASSOPTIONcompsoc
  \section*{Acknowledgments}
\else
  \section*{Acknowledgment}
\fi

The authors would like to extend their sincere thanks to Hyeonwoo Yu and Sanghyun Cho for valuable discussions and suggestions.
\vspace{-0.12in}

\ifCLASSOPTIONcaptionsoff
  \newpage
\fi



\newpage
\begin{IEEEbiography}[{\includegraphics[width=1in,height=1.25in,clip,keepaspectratio]{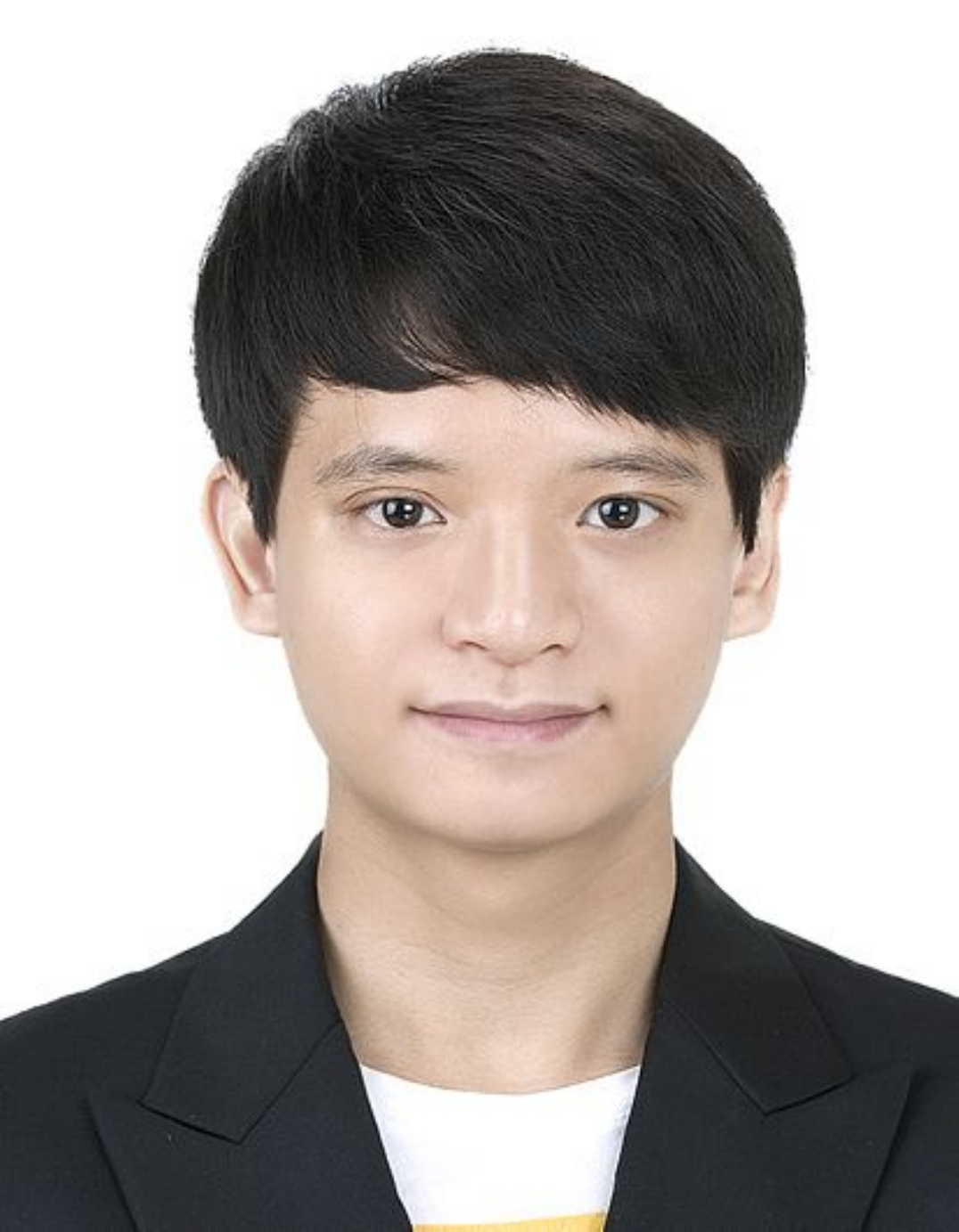}}]{Jihoon Moon}
received the B.S. and M.S. degrees in Electrical and Computer Engineering 
from the Seoul National University, South Korea, in 2013 and 2015, respectively.
His current research focuses on online domain adaptation, computer vision and robotics. 
Currently he is pursuing the Ph.D. degree at the Elmore Family 
School of Electrical and Computer Engineering, Purdue University, West Lafayette, Indiana, U.S.A.
\end{IEEEbiography}

\vskip 0pt plus -1fil

\begin{IEEEbiography}[{\includegraphics[width=1in,height=1.25in,clip,keepaspectratio]{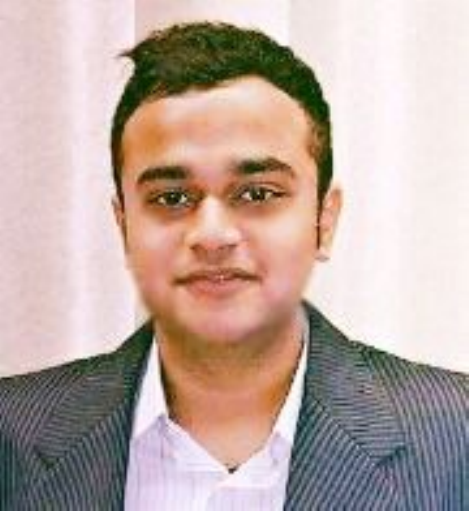}}]{Debasmit Das} (S'11-M'21)
received the B.Tech. degree in Electrical Engineering from the Indian Institute of Technology, Roorkee in 2014 and the Ph.D. degree from the School of Electrical and Computer Engineering, Purdue University, West Lafayette, Indiana in 2020. He is currently a Senior Machine Learning Researcher at Qualcomm AI Research, Qualcomm Technologies, Inc., San Diego, CA, investigating efficient and personalized learning solutions. He is an associate editor of Wiley Applied AI Letters and he also regularly reviews machine learning articles for IEEE, ACM, Springer and Elesevier journals.
\end{IEEEbiography}

\vskip 0pt plus -1fil

\begin{IEEEbiography}[{\includegraphics[width=1in,height=1.25in,clip,keepaspectratio]{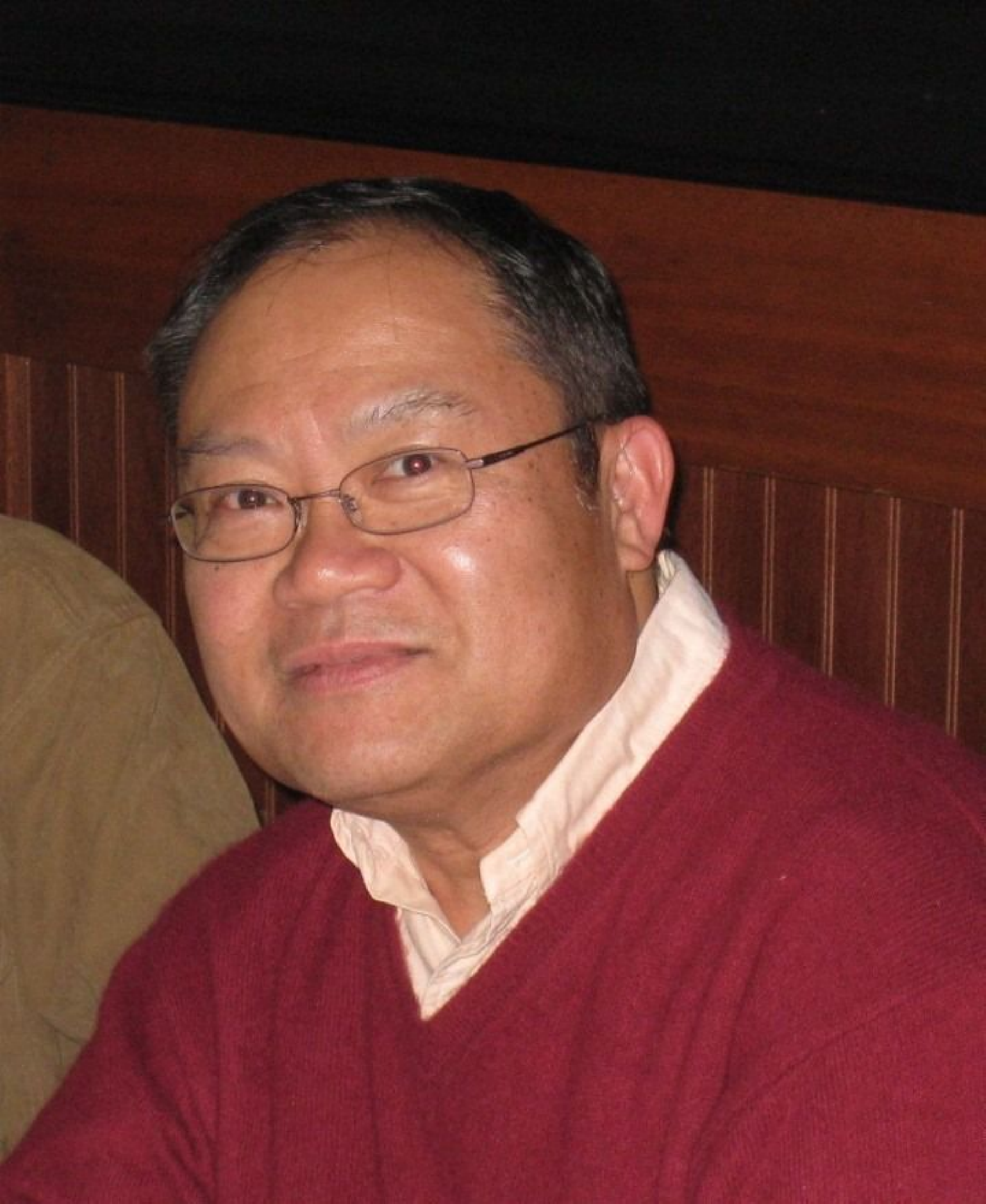}}]{C. S. George Lee}
(S'71--M'78--SM'86--F'93)
is a Professor of Electrical and
Computer Engineering at Purdue University, West
Lafayette, Indiana. His current research focuses on
transfer learning and skill learning, human-centered
robotics, and neuro-fuzzy systems. 
In addition to publishing extensively in those areas, 
he has co-authored two graduate textbooks,
Robotics: Control, Sensing, Vision, and Intelligence
(McGraw-Hill, 1986) and Neural Fuzzy Systems:
A Neuro-Fuzzy Synergism to Intelligent Systems
(Prentice-Hall, 1996).
Dr. Lee is an IEEE Fellow, a recipient of the IEEE Third Millennium
Medal Award, the Saridis Leadership Award and the Distinguished Service
Award from the IEEE Robotics and Automation Society. 
Dr. Lee received the Ph.D. degree from Purdue University, West Lafayette, Indiana, 
and the M.S.E.E. and B.S.E.E. degrees from Washington State University, Pullman, Washington. 
\end{IEEEbiography}

\end{document}